\documentclass[lettersize,journal]{IEEEtran}
\usepackage{amsmath,amsfonts}
\usepackage{algorithmic}
\usepackage{algorithm}
\usepackage{array}
\usepackage[caption=false,font=normalsize,labelfont=sf,textfont=sf]{subfig}
\usepackage{textcomp}
\usepackage{stfloats}
\usepackage{url}
\usepackage{verbatim}
\usepackage{graphicx}
\usepackage{cite}
\usepackage{soul}
\usepackage[utf8]{inputenc} 
\usepackage[T1]{fontenc}    
\usepackage{hyperref}       
\usepackage{url}            
\usepackage{booktabs}       
\usepackage{amsfonts}       
\usepackage{nicefrac}       
\usepackage{microtype}      
\usepackage{xcolor}         
\usepackage{multirow}
\usepackage{colortbl}
\usepackage{booktabs}
\usepackage{graphicx}
\usepackage{amsmath}
\usepackage{bm}
\usepackage{amssymb}
\usepackage{graphicx,wrapfig,lipsum}
\usepackage{caption}
\usepackage{tabularx}
\usepackage{bbding}

\begin{document}
\title{FreeSplat++: Generalizable 3D Gaussian Splatting for Efficient Indoor Scene Reconstruction}

\author{\textbf{Yunsong Wang, Tianxin Huang, Hanlin Chen, Gim Hee Lee}
\\{School of Computing, National University of Singapore}
\\
\texttt{yunsong@comp.nus.edu.sg} \quad
    \texttt{gimhee.lee@nus.edu.sg}
\\{\ttfamily \url{https://wangys16.github.io/FreeSplatPP-Page}}}





\twocolumn[{%
\vspace{-0.3cm}
\renewcommand\twocolumn[1][]{#1}%
\maketitle
\begin{center}
    \centering
    \captionsetup{type=figure}
    \includegraphics[width=1\textwidth]{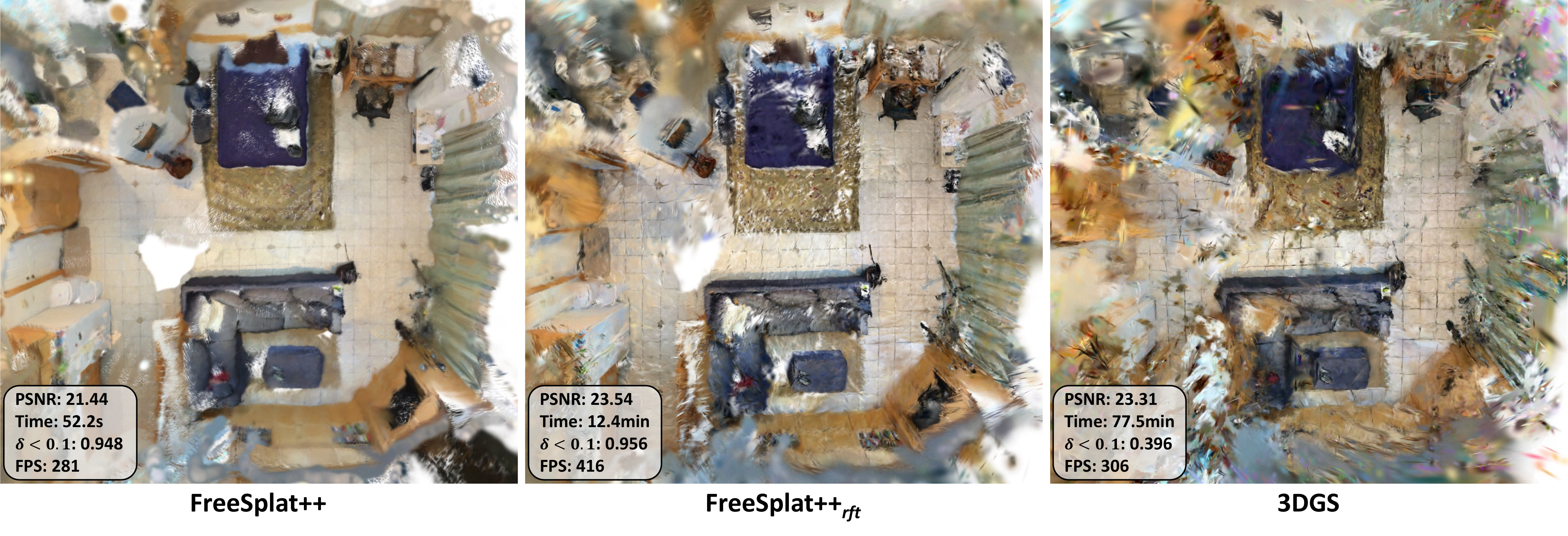}
    \captionof{figure}{\textbf{Results on whole scene reconstruction.} FreeSplat++$_{rft}$ is the per-scene fine-tuned results. Our model excels in efficiently reconstructing geometrically accurate 3D Gaussian primitives. Furthermore, FreeSplat++ shows superior view consistency, \textit{e.g.} when rendered from Bird-Eye's View, demonstrating the significance of generalizable 3DGS for whole scene reconstruction.}
    \label{fig:teaser}
\end{center}%
}]

\begin{abstract}
  Recently, the integration of the efficient feed-forward scheme into 3D Gaussian Splatting (3DGS) has been actively explored. However, most existing methods focus on sparse view reconstruction of small regions and cannot produce eligible whole-scene reconstruction results in terms of either quality or efficiency. In this paper, we propose FreeSplat++, which focuses on extending the generalizable 3DGS to become an alternative approach to large-scale indoor whole-scene reconstruction, which has the potential of significantly accelerating the reconstruction speed and improving the geometric accuracy. 
  To facilitate whole-scene reconstruction, we initially propose the Low-cost Cross-View Aggregation framework to efficiently process extremely long input sequences.
  Subsequently, we introduce a carefully designed pixel-wise triplet fusion method to incrementally aggregate the overlapping 3D Gaussian primitives from multiple views, adaptively reducing their redundancy. 
  Furthermore, given the fused 3DGS primitives with accumulated weights after the fusion step, we propose a weighted floater removal strategy that can effectively reduce floaters, which serves as an explicit depth fusion approach that is tailored for generalizable 3DGS methods and becomes crucial in whole-scene reconstruction.  
  After the feed-forward reconstruction of 3DGS primitives, we investigate a depth-regularized per-scene fine-tuning process. Leveraging the dense, multi-view consistent depth maps obtained during the feed-forward prediction phase for an extra constraint, we refine the entire scene’s 3DGS primitive to enhance rendering quality while preserving geometric accuracy.
  Extensive experiments confirm that our FreeSplat++ significantly outperforms existing generalizable 3DGS methods, especially in whole scene reconstructions. Compared to conventional per-scene optimized 3DGS approaches, our method with  depth-regularized per-scene fine-tuning demonstrates substantial improvements in reconstruction accuracy and a notable reduction in training time.
  
\end{abstract}

\section{Introduction}

In recent years, differentiable neural rendering has attracted considerable interest for applications in novel view synthesis \cite{nerf, ibrnet, mvsnerf, mipnerf360, ngp, 3dgs, mipsplatting, 4dgs}, mesh reconstruction \cite{neus, neuralangelo, neusg, vcr}, scene understanding \cite{nesf, semanticray, gov}, and embodied AI \cite{embodied1, embodied2}. Among these techniques, Neural Radiance Fields (NeRF) \cite{nerf, mipnerf360} is a milestone for learning the geometry and appearance of 3D scenes through an implicit radiance field. However, NeRF is computationally expensive in both training and rendering phases. 
An alternative approach, 3D Gaussian Splatting (3DGS) \cite{3dgs, mipsplatting}, represents scenes explicitly using a set of anisotropic 3D Gaussian primitives and optimizes them via differentiable  rendering. This method has gained increasing popularity for its ability to produce high-quality renderings in real time. However, despite these advancements, vanilla 3DGS still requires tens of minutes for per-scene optimization, limiting its broader applicability.

\begin{figure}[t]
    \centering
    \includegraphics[width=1\linewidth]{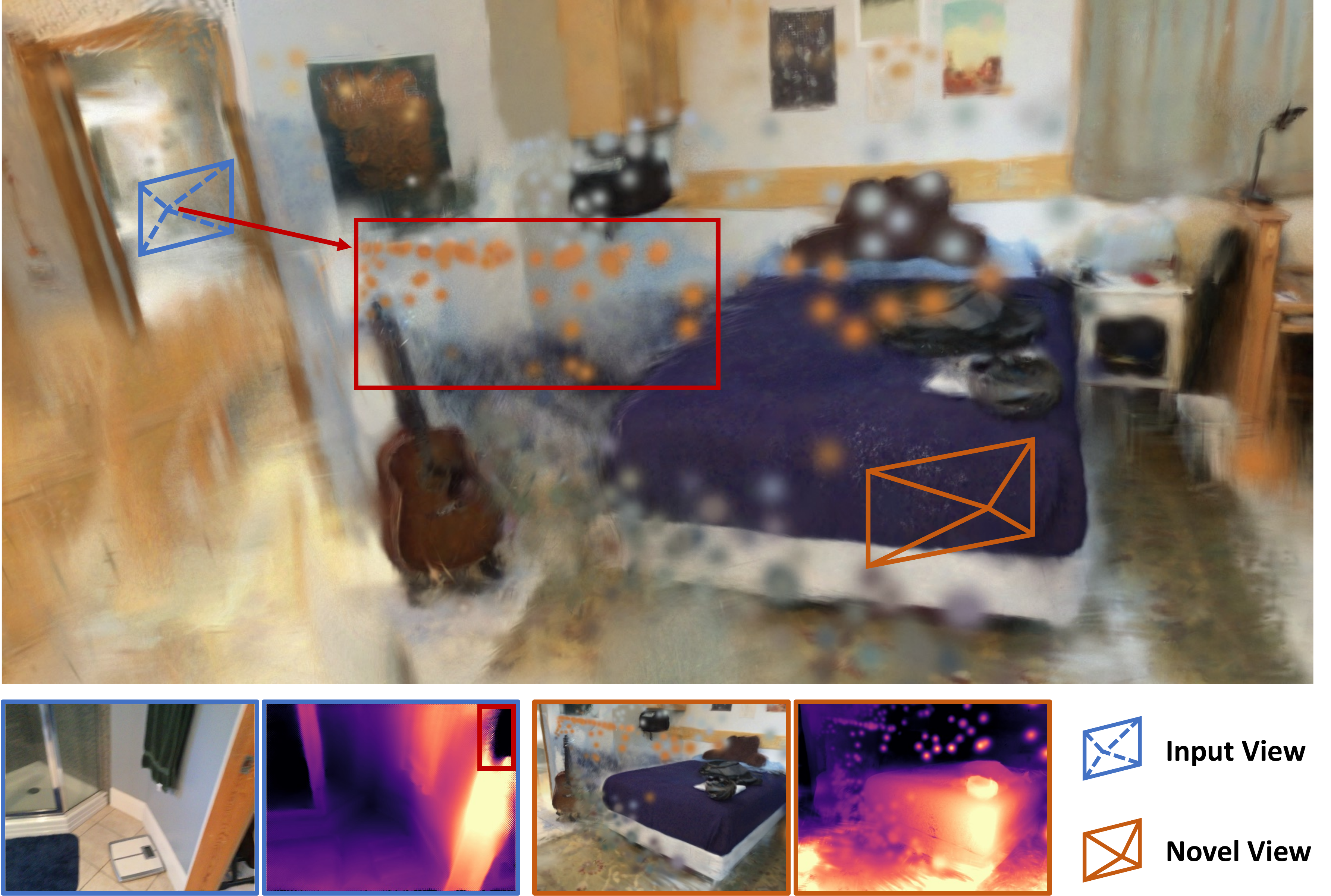}
    \caption{\textbf{Visualization of the floaters challenge in generalizable 3DGS.} The noise of predicted depth maps (\textcolor{red}{red} regions) may lead to severe floaters in whole scene reconstruction.} 
    \label{fig:floaters}
\end{figure}

To this end, there have been multiple attempts \cite{pixelsplat, mvsplat, gps, latentsplat} to transfer the vanilla 3DGS to a feed-forward manner. 
However, most of them solely focus on sparse view interpolation within a considerable narrow view range, making them less effective than vanilla 3DGS for explicitly reconstructing large-scale 3D scenes. This limitation undermines the full potential of 3DGS as an explicit representation. 
More recently, FreeSplat \cite{freesplat} has explored extending generalizable 3DGS for reconstructing large-scale 3D scenes. It introduces an efficient backbone capable of processing long-sequence inputs and a Gaussian fusion mechanism that incrementally fuses global Gaussian triplets with local triplets from input views, effectively reducing redundancy in overlapping regions. FreeSplat surpasses previous baselines in geometric accuracy, rendering quality, and efficiency, particularly in long-sequence reconstruction.
Despite these advancements, FreeSplat still faces challenges in reconstructing whole indoor scenes \textbf{because}: when given input images spanning the whole 3D scene, its fusion module cannot remove the floaters largely deviating from the surfaces.
In these cases, the number of Gaussian primitives becomes enormous and undermines real-time rendering, and the rendering quality of the overlapping regions is also degraded.
A visualization of the floaters challenge of FreeSplat is shown in Figure \ref{fig:floaters}, in which the model struggles to deal with the noisy depth map from some specific input views, and leaves severe floaters in the whole-scene reconstruction. 
The main reason of such challenge is the lack of operations to check the multi-view depth consistency and remove the inconsistent regions.
However, existing multi-view depth consistency checking methods, \textit{e.g.} in \cite{mvsgaussian}, produces holes in the final reconstruction and cannot be directly integrated to feed-forward 3DGS.

Therefore, we introduce FreeSplat++, an enhanced generalizable 3DGS framework specifically designed for feed-forward whole-scene reconstruction that achieves high rendering quality, reconstruction accuracy, and efficiency. First, we propose a low-cost CNN-based framework to efficiently aggregate multi-view images through cost volumes, enabling efficient management of whole-scene information. Secondly, drawing inspiration from traditional TSDF Fusion techniques \cite{volumetric, kinectfusion}, we carefully design an improved Pixel-wise Triplet Fusion (PTF) mechanism to fuse pixel-wise Gaussian primitives under current view, with the closest global Gaussian primitives within a broader range, in order to further reduce the redundancy of 3DGS primitives and remove the floaters during fusion. 
Additionally, inspired by the voxel-based averaging of TSDF values in TSDF Fusion — which effectively mitigates noise in depth maps — we introduce a Weighted Floater Removal (WFR) strategy. This strategy incrementally projects the global 3DGS onto the input views, and adjusts the Gaussian opacities according to predicted/rendered depth maps and the Gaussian weights accumulated in the PTF process. 
Subsequently, to further improve the rendering quality while maintaining the geometric accuracy, we introduce a depth-regularized per-scene fine-tuning step to efficiently optimize the feed-forward predicted 3DGS.
According to our experiments, our improved feed-forward pipeline can effectively remove floaters and efficiently reconstruct complex large 3D scenes, and our per-scene fine-tuned results show significant advantages over vanilla 3DGS in terms of both reconstruction accuracy and efficiency. 

Overall, our \textbf{contributions} can be summarized as:

\begin{enumerate}
    \item This work represents a pioneering effort in generalizable 3D Gaussian Splatting (3DGS) for large scale reconstruction. To the best of our knowledge, it is the first framework to effectively apply generalizable 3DGS for indoor whole-scene reconstruction.
    \item To address the challenge of feed-forward whole scene reconstruction, we improve the fusion mechanism in FreeSplat \cite{freesplat}, and propose an effective weighted floater removal strategy that can be seamlessly integrated into the framework to further remove the floaters.
    \item We introduce a depth-regularized per-scene fine-tuning step after feed-forward 3DGS prediction. Leveraging the efficient and geometrically accurate initialization of 3DGS from our feed-forward pipeline, our whole-scene reconstruction framework provides a competitive alternative to vanilla 3DGS, achieving a significant improvement on reconstruction accuracy within considerably shorter training time.
\end{enumerate}

\section{Related Work}

\noindent\textbf{Novel View Synthesis.} Novel View Synthesis aims to generate images from unseen views that are consistent with the given multi-view images. Traditional attempts in novel view synthesis mainly employed voxel grids \cite{volume1, volume2} or multi-plane images \cite{multiplane}. Neural Radiance Fields (NeRF) \cite{nerf} was a milestone in novel view synthesis which learns 3D geometry and radiance purely from color images. It represents the 3D scene using an implicit radiance field, such that the novel view renderings can be conducted through a differentiable volume rendering mechanism. However, a major drawback of NeRF is its slow rendering speed, attributable to the computationally intensive ray-based volume rendering of numerous 3D points. Recently, 3D Gaussian Splatting (3DGS) has been proposed as a real-time method for novel view synthesis, which explicitly models the 3D scene as sets of anisotropic 3D Gaussians with learnable parameters, such that they can optimize the 3DGS parameters while performing adaptive densify control to fit to the given set of training images.  A crucial advantage of 3DGS is its efficient tile-based rendering process, which significantly outperforms NeRF in rendering speed, achieving real-time performance. Despite the high rendering quality and real-time rendering speed, 3DGS still requires tens of minutes for per-scene optimization, and is prone to overfitting to the training views especially in complex indoor 3D scenes. 
This tendency results in producing high-quality images from training views while often failing to reconstruct accurate 3D scene geometry due to lack of depth regularization. This issue has been highlighted in recent studies \cite{geogaussian, drgs}, where it has been shown that such overfitting leads to substantial performance degradation when rendering from extrapolated views.

\noindent\textbf{Generalizable Novel View Synthesis.} One bottleneck of the traditional NeRF-based and 3DGS-based methods is the requirement of per-scene optimization instead of direct feeding-forward processing. To circumvent the expensive optimization of NeRF, several studies \cite{pixelnerf, mvsnerf, wang2021ibrnet, regnerf, pointnerf, enerf} have focused on learning effective priors to predict 3D geometry from images in a feed-forward manner. Commonly, these methods project ray-marching sampled points onto given source views to aggregate multi-view features, conditioning the prediction of the implicit fields on source views instead of 3D point coordinates. Despite their high rendering quality, they still suffer from relatively slow rendering speed, and requires separate feed-forward process when rendering from different target views, limiting their widespread application. Recently, efforts have been made towards developing a generalizable 3D Gaussian Splatting framework \cite{pixelsplat, mvsplat, latentsplat, gps, transplat, hisplat} that enables feed-forward 3DGS reconstruction from sparse-view images. The common pipeline is to predict pixel-aligned Gaussian parameters and depths for each input view, and unproject the Gaussians into the 3D space. During training, the model is supervised by the renderings from the interpolated views, such that the model can learn 3D geometry purely from color images. However, these methods generally focus on very sparse view cases ($\leq 3$ views), and the rendering quality and efficiency degenerates when the number of input views increases, as shown in \cite{pixelsplat, mvsplat, freesplat}, limiting their application in large 3D scene reconstruction. Some works \cite{gps, ggrt} only focus on view interpolation between two input views and conduct separate feed-forward process as NeRF-based methods, which greatly undermines the explicit representation potential of 3DGS and can only handle view interpolation. Recently, the prior version of this work, FreeSplat \cite{freesplat}, proposes a fusion mechanism to aggregate the overlapping 3D gaussians for higher rendering quality and efficiency, and its transformer-free pipeline supports training and inference on longer input sequence, and can effectively aggregate more nearby views to learn accurate 3D geometry. Nonetheless, this fusion technique struggles to eliminate floaters that deviate significantly from surfaces, and its rendering quality and efficiency greatly degrades in whole scene reconstructions.

\vspace{3pt}\noindent\textbf{Indoor Scene Reconstruction.} Efforts in indoor scene reconstruction vary widely in technique and focus. One line of efforts in indoor scene reconstruction focuses on extracting 3D mesh using voxel volumes \cite{con, neuralrecon, vortx} and TSDF-fusion \cite{simplerecon}, while these mesh-based methods generally produces color renderings of inferior quality compared to those achieved by neural rendering-based methods. On the other hand, the SLAM-based methods \cite{niceslam, gsslam, splatam} normally require dense sequence of RGB-D input and per-scene tracking and mapping to achieve satisfying pose estimation and reconstruction accuracy in complex scenes. Another paradigm of 3D reconstruction \cite{volsdf, monosdf, neuralangelo} learns implicit Signed Distance Fields from RGB input, while demanding intensive per-scene optimization. Another recent work SurfelNeRF \cite{gao2023surfelnerf} learns a feed-forward framework to map a sequence of images to 3D surfels which support photorealistic image rendering, while they do not integrate multi-view features and rely on external depth estimator or ground truth depth maps. In contrary, we propose an end-to-end model that can work without ground truth depth map input or supervision, enabling accurate localization of 3D Gaussians using only photometric losses. Furthermore, recent works utilizing 3DGS for indoor scene reconstruction \cite{gaussianroom, 2dgsroom, vcr} regularize 3DGS localization during per-scene optimization. However, these methods are generally inefficient due to their more costly per-scene training (\textit{e.g.} predicting monocular cues, jointly training an SDF implicit field), and the monocular cues can lead to suboptimal performance due to their multi-view inconsistency (\textit{c.f.} Table \ref{tab:whole_scannet}, \ref{tab:whole_scannetpp}, Figure \ref{fig:whole_gs}). Instead, this work provides a novel approach to whole scene reconstruction through extending the generalizable 3DGS method for higher efficiency and more accurate geometry.

\section{Preliminary}
\noindent\textbf{Vanilla 3DGS.} 3D-GS \cite{3dgs} explicitly represents a 3D scene with a set of Gaussian primitives which are parameterized via a 3D covariance matrix $\bm{\Sigma}$ and mean $\bm{\mu}$:
\begin{equation}
    G(\textbf{p})=\mathrm{exp}(-\frac{1}{2}\left(\bm{p}-\bm{\mu}\right)^\top\bm{\Sigma}^{-1}\left(\bm{p}-\bm{\mu}\right)),
\end{equation}
where $\bm{\Sigma}$ is decomposed into $\bm{\Sigma}=\bm{\mathrm{R}}\bm{\mathrm{S}}\bm{\mathrm{S}}^\top\bm{\mathrm{R}}^\top$ using a scaling matrix $\bm{\mathrm{S}}$ and a rotation matrix $\bm{\mathrm{R}}$ to maintain positive semi-definiteness. During rendering, the 3D Gaussian is transformed into the image coordinates with world-to-camera transform matrix $\bm{\mathrm{W}}$ and projected onto image plane with projection matrix $\bm{\mathrm{J}}$, and the 2D covariance matrix $\bm{\mathrm{\Sigma}}'$ is computed as $\bm{\Sigma}'=\bm{\mathrm{J}}\bm{\mathrm{W}}\bm{\mathrm{\Sigma}}\bm{\mathrm{W}}^\top\bm{\mathrm{J}}^\top$. 
We then obtain a 2D Gaussian $G^{2D}$ with the covariance $\bm{\Sigma}'$ in 2D, and the color rendering is computed using point-based alpha-blending on each ray:
\begin{equation}
    \bm{\mathrm{C}}(\bm{\mathrm{x}})=\sum_{i\in N}\bm{\mathrm{c}}_i \alpha_i G^{2D}_i(\bm{\mathrm{x}})\prod_{j=1}^{i-1}(1-\alpha_j G^{2D}_j(\bm{\mathrm{x}})),
    \label{eq:render}
\end{equation}
where $N$ is the number of Gaussian primitives, $\alpha_i$ is a learnable opacity, and $\bm{\mathrm{c}}_i$ is view-dependent color defined by spherical harmonics (SH) coefficients $\bm{\mathrm{s}}$. The Gaussian parameters are optimized by a photometric loss to minimize the difference between renderings and image observations.

\vspace{3pt}\noindent\textbf{Generalizable 3DGS.} Unlike vanilla 3DGS that optimizes per-scene Gaussian primitives, recent generalizable 3DGS \cite{pixelsplat, gps} predict pixel-aligned Gaussian primitives $\{\bm{\Sigma},\bm{\alpha}, \bm{\mathrm{s}}\}$ and depths $\bm{d}$, such that the pixel-aligned Gaussian primitives can be unprojected to 3D coordinates $\bm{\mu}$. The Gaussian parameters are predicted by 2D encoders, which are optimized by the photometric loss through rendering from novel views. However, existing methods mostly focus on view interpolation within narrow view range, and struggle to extend to long sequence input or whole scene reconstruction due to severe floaters and extremely large number of gaussians. To this end, we propose FreeSplat++ as a prior attempt to accomplish the challenging task of feed-forward whole scene reconstruction with accurately localized 3D Gaussians.

\section{Our Methodology}
\label{sec:method}

\begin{figure*}[t!]
    \centering
    \includegraphics[width=1\textwidth]{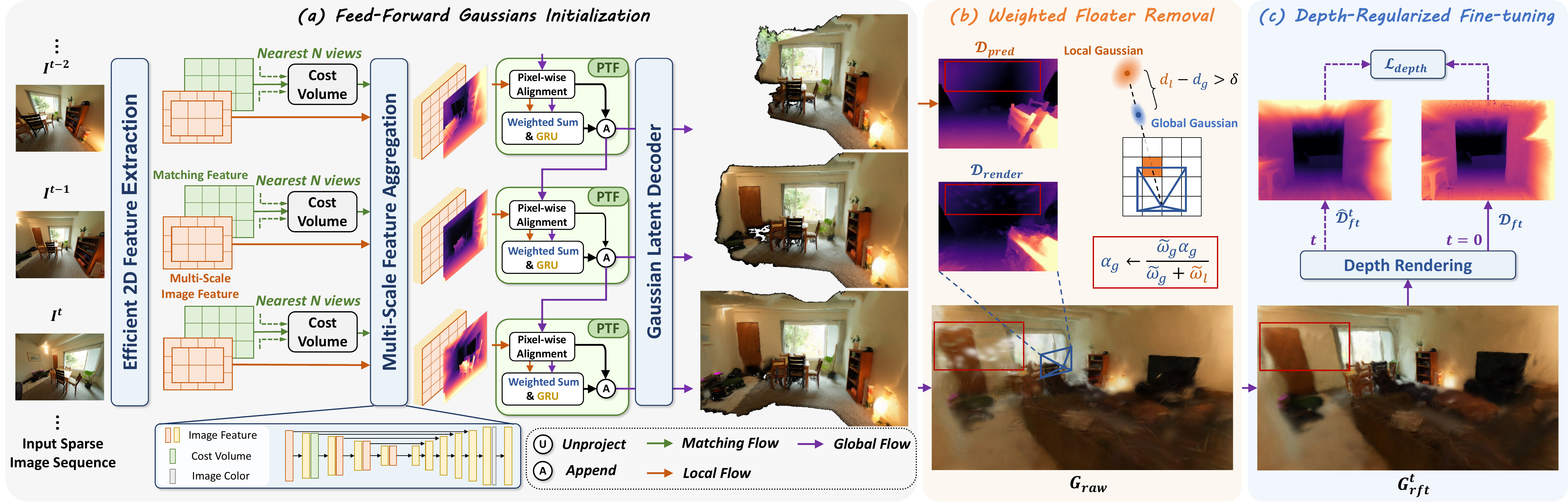}
    \caption{\textbf{Framework of FreeSplat++.} The high-level design of FreeSplat++ includes: \textbf{(a) Feed-Forward Gaussians Initialization}: given input sparse sequence of images, we construct cost volumes between nearby views and introduce Pixel-aligned Triplet Fusion (PTF) module, where we progressively aggregate and update local/global Gaussian triplets based on pixel-wise alignment. \textbf{(b) Weighted Floater Removal}: Leverage the accumulated gaussian weights in our PTF process, we further align the global and local gaussians and incrementally adjust the gaussian opacities. \textbf{(c) Depth-Regularized Fine-tuning}: We can optionally conduct a fast per-scene fine-tuning step with multi-view consistent depth regularization thanks to our geometrically accurate gaussian initialization.}
    \label{fig:framework}
\end{figure*}
\subsection{Overview}
The overview of our method is illustrated in Figure \ref{fig:framework}. Given a sparse sequence of RGB images, we propose a CNN-based framework in which we build cost volumes adaptively between nearby views, and predict depth maps to unproject the 2D feature maps into 3D Gaussian triplets. 
Then, inspired by the traditional TSDF Fusion methods, we propose the Pixel-aligned Triplet Fusion (PTF) module to progressively align the global with the local Gaussian triplets, such that we can fuse the redundant 3D Gaussians in the latent feature space and aggregate cross-view Gaussian features before decoding, in order to mitigate redundancy and aggregate point-level latent features across multi views. Furthermore, we propose a Weighted Floater Removal (WFR) module to further remove floaters, in which we leverage the accumulated weights in the PTF process and adjust the opacities of the floater gaussians. Subsequently, we investigate a depth-regularized per-scene fine-tuning step to further improve the rendering quality while preserving the geometry accuracy of the feed-forward predicted 3DGS. Our method is capable of taking in whole scene image sequence and efficiently reconstructing whole scene 3DGS with high geometry accuracy.

\subsection{Low-cost Cross-View Aggregation}
\label{sec:cost_volume}
\noindent\textbf{Efficient 2D Feature Extraction.} Given a sparse sequence of posed images $\{\bm{I}^t\}_{t=1}^T$, we first feed them into a shared 2D backbone to extract multi-scale embeddings $\bm{F}_e^t$ and matching feature $\bm{F}_m^t$. 
Unlike \cite{pixelsplat, mvsplat} which rely on patch-wise transformer-based backbones \cite{vit, swin} that can lead to quadratically expensive computations, we leverage pure CNN-based backbones \cite{efficientnet, resnet} to efficiently handle long sequence of high-resolution inputs.

\vspace{3pt}\noindent\textbf{Adaptive Cost Volume Formulation.} To explicitly integrate camera pose information given arbitrary length of input images, we propose to adaptively build cost volumes between nearby views. For current view $\bm{I}^t$ with pose $\bm{P}^t$ and matching feature $\bm{F}_m^t\in\mathbb{R}^{C_m\times\frac{H}{4}\times\frac{W}{4}}$, we adaptively select its $N$ nearby views $\{\bm{I}^{t_n}\}_{n=1}^{N}$ with poses $\{\bm{P}^{t_n}\}_{n=1}^{N}$  based on pose 
proximity, and construct cost volume via plane sweep stereo \cite{sweep1, sweep2}. Specifically, we define a set of $K$ virtual depth planes $\{d_k\}_{k=1}^{K}$ that are uniformly spaced within $[d_{near}, d_{far}]$, and warp the nearby view features to each depth plane $d_k$ of current view:
\begin{equation}
    \Tilde{\bm{F}}_m^{t_n,k}=\mathrm{Trans}(\textbf{P}^{t_n},\textbf{P}^{t})\bm{F}_m^{t_n},
\end{equation}
where $\mathrm{Trans}(\textbf{P}^{t_n},\textbf{P}^{t})$ is the transformation matrix from view $t_n$ to $t$. 
The cost volume $\bm{F}_{\mathrm{cv}}^t\in \mathbb{R}^{K\times\frac{H}{4}\times\frac{W}{4}}$ is then defined as:
\begin{equation}
    \bm{F}_{\rm{cv}}^t(k)=f_\theta\left((\frac{1}{N}\sum_{n=1}^N\mathrm{cos}(\bm{F}_m^t,\Tilde{\bm{F}}_m^{t_n,k}))\oplus(\frac{1}{N}\sum_{n=1}^{N}\Tilde{\bm{F}}_m^{t_n,k})\right),
\end{equation}
where $\bm{F}_{\rm{cv}}^t[k]$ is the $k$-th dimension of $\bm{F}_{\rm{cv}}^t$, $\mathrm{cos}(\cdot)$ is the cosine similarity, $\oplus$ is feature-wise concatenation, and $f_{\theta}(\cdot)$ is a $1\times1\;\mathrm{CNN}$ mapping to dimension of $1$. The concatenation of the averaged multi-view features is to embed more abundant information from nearby views.

\vspace{3pt}\noindent\textbf{Multi-Scale Feature Aggregation.}
\label{decoding}
The embedding of the cost volume plays a significant part to accurately localize the 3D Gaussians. To this end, inspired by existing depth estimation methods \cite{deepvideomvs, simplerecon}, we design a multi-scale encoder-decoder structure, such that to fuse multi-scale image features with the cost volume and propagate the cost volume information to broader receptive fields. 
Specifically, the multi-scale encoder takes in $\bm{F}_{\mathrm{cv}}^t$ and the output is concatenated with $\{\bm{F}_s^t\}$. We then send it into a UNet++ \cite{unet++}-like decoder to upsample to \textit{half} of full resolution and predict a depth candidates map $\bm{D}_c^t\in\mathbb{R}^{K\times \frac{H}{2}\times \frac{W}{2}}$ and Gaussian triplet map $\bm{F}^t_l\in\mathbb{R}^{C\times \frac{H}{2}\times \frac{W}{2}}$, since we empirically find that lowering the resolution of the unprojection map to half leads to similar performance with much higher efficiency, which becomes indispensable in whole scene reconstruction to keep the number of gaussians affordable.

We then predict the depth map through soft-argmax to bound the depth prediction between near and far:
\begin{equation}
    \label{eq:depth}
    \bm{D}^t=\sum_{k=1}^K\mathrm{softmax}(\bm{D}_c^t)_k\cdot d_k.
\end{equation}
Finally, the pixel-aligned Gaussian triplet map $\bm{F}^t_l$ is unprojected to 3D Gaussian triplet $\{\bm{\mu}_l^t,\bm{\omega}_l^t,\bm{f}^t_l\}$, where $\bm{\mu}_l^t\in\mathbb{R}^{3\times \frac{HW}{4}}$ are the Gaussian centers, $\bm{\omega}_l^t\in\mathbb{R}^{1\times \frac{HW}{4}}$ are weights between $(0,1)$, and $\bm{f}^t_l\in\mathbb{R}^{(C-1)\times \frac{HW}{4}}$ are Gaussian triplet features.  
\begin{figure}[t]
    \centering
    \includegraphics[width=1\linewidth]{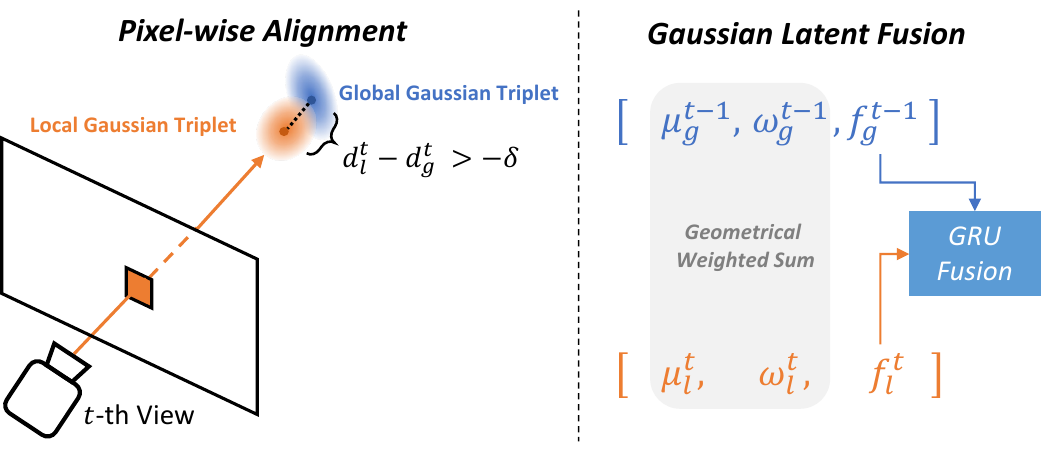}
    \caption{\textbf{Visual illustration of PTF.} The PTF incrementally projects current global Gaussians to input views and 
    computes their pixel-wise distance with local Gaussians. 
    Nearby local Gaussians are then fused using a lightweight Gate Recurrent Unit (GRU) network \cite{gru}.} 
    \label{fig:illustration}
\end{figure}
\subsection{Pixel-wise Triplet Fusion}
\label{sec:plf}
One limitation of previous generalizable 3DGS methods is the redundancy of Gaussians, especially in whole scene reconstruction. Since 
we need multi-view observations to predict accurately localized 3D Gaussians in indoor scenes, the pixel-aligned 3D Gaussians become redundant in frequently observed regions. Furthermore, previous methods integrate multi-view Gaussians of the same region simply through their opacities, leading to suboptimal performance due to lack of post aggregation, and extremely large number of gaussians in whole scene reconstruction (\textit{cf.} Table \ref{tab:ablation}). Consequently, inspired by previous methods \cite{neuralrecon, gao2023surfelnerf}, we propose the Pixel-wise Triplet Fusion (PTF) module 
which can significantly remove redundant Gaussians in the overlapping regions and explicitly aggregate multi-view observation features in the latent space. Specifically, during the unprojection of Gaussians, we incrementally align the global Gaussians with local ones in each view using Pixel-wise Alignment to select the redundant pairs, and aggregate their latent features through GRU network.

\vspace{3pt}\noindent\textbf{Pixel-wise Alignment.} Given the Gaussian triplets $\{\bm{\mu}^t_l, \bm{\omega}^t_l, \bm{f}_l^t\}_{t=1}^T$, we start from $t=1$ where the global Gaussian triplet is empty. In the $t$-th step, we have global 3D Gaussian triplets $\{\bm{\mu}^{t-1}_l, \bm{\omega}^{t-1}_l, \bm{f}_l^{t-1}\}$ which were unprojected from previous views. We first project the global Gaussian triplet centers $\bm{\mu}_g^{t-1}$ onto the $t$-th view:
\begin{equation}
    \bm{\mathrm{p}}_g^t:=\{\bm{\mathrm{x}}_g^t, \bm{\mathrm{y}}_g^t, \bm{\mathrm{d}}_g^t\} = \bm{P}^t\bm{\mu}_g^{t-1},
\end{equation}
where $[\bm{\mathrm{x}}_g^t, \bm{\mathrm{y}}_g^t, \bm{\mathrm{d}}_g^t]\in\mathbb{R}^{3\times M}$ are the projected 2D coordinates and corresponding depths. We then correspond the local Gaussian triplets with the pixel-wise nearest projections within a threshold. Specifically, for the $i$-th pixel under current view, we have its local Gaussian with 2D coordinate $[\bm{\mathrm{x}}_l^t(i), \bm{\mathrm{y}}_l^t(i)]$ and depth $\bm{d}_l^t(j)$. We first find its intra-pixel global projection set $\bm{\mathcal{S}}_i$: 
\begin{equation}
\label{eq:projection}
    \bm{\mathcal{S}}_i^t := \{j\mid [\bm{\mathrm{x}}_g^t(j)]=\bm{\mathrm{x}}_l^t(i), [\bm{\mathrm{y}}_g^t(j)]=\bm{\mathrm{y}}_l^t(i)\},
\end{equation}
where $[\,\cdot\,]$ is the rounding operator. 
Subsequently, we search for valid correspondence with minimum depth difference under a threshold:
\begin{equation}
\label{eq:alignment}
    m_i=\begin{cases}
    \begin{array}{c@{\quad}l}
\mathop{\arg\min}\limits_{j\in\bm{\mathcal{S}}_i^t}\bm{\mathrm{d}}^t_g(j), &\text{if}\;\bm{\mathrm{d}}_l^t(j)-\mathop{\min}\limits_{j\in\bm{\mathcal{S}}_i^t}\bm{\mathrm{d}}^t_g(j)>-\delta\\
\hfil \varnothing, &\text{otherwise}
\end{array},
\end{cases}
\end{equation}
where $\delta$ is a threshold. Inspired by traditional TSDF Fusion, we expand the depth range of fusion comparing to the condition in FreeSplat \cite{freesplat}, in order to remove more foreground floaters in more complex whole scene reconstruction scenarios. We define the valid correspondence set as:
\begin{equation}
    \bm{\mathcal{F}}^t:=\{(i,m_i)\mid i=1,...,HW;\;m_i\neq\varnothing\}.
\end{equation}

\vspace{3pt}\noindent\textbf{Gaussian Triplet Fusion.} 
After the pixel-wise alignment, we remove the redundant 3D Gaussians through merging the validly aligned triplet pairs. Given a pair $(i,m_i)\in\rm{\mathcal{F}}^t$, 
we compute the weighted sum of their center coordinates and sum their weights to restrict the 3D Gaussian centers to lie between the triplet pair: 
\begin{gather}
\bm{\mu}^t_g(m_i)\leftarrow\frac{\bm{\omega}^t_l(i)\bm{\mu}^t_l(i)+\bm{\omega}^{t-1}_g(m_i)\bm{\mu}^{t-1}_g(m_i)}{\bm{\omega}^t_l(i)+\bm{\omega}^{t-1}_g(m_i)},\\
\bm{\omega}^t_g(m_i)\leftarrow\bm{\omega}^t_l(i)+\bm{\omega}^{t-1}_g(m_i).
\end{gather}

We then aggregate the aligned local and global Gaussian latent features through a lightweight GRU network:
\begin{equation}
    \bm{f}_g^t(m_i)=\operatorname{GRU}(\bm{f}_l^t(i), \bm{f}_g^{t-1}(m_i)).
\end{equation}
Then we append the fused Gaussian triplets with the other unaligned local and global Gaussian triplets, forming new global triplets $\{\bm{\mu}^t_l, \bm{\omega}^t_l, \bm{f}_l^t\}$.

\vspace{3pt}\noindent\textbf{Gaussian primitives decoding.} After the Pixel-wise Triplet Fusion, we can decode the global Gaussian triplets into Gaussian primitives:
\begin{equation}
    \bm{\Sigma},\bm{\alpha},\bm{\mathrm{s}}=\operatorname{MLP}_d(\bm{f}_g^T)
\end{equation}
and 
Gaussian centers $\bm{\mu}=\bm{\mu}_g^\top$. 
Our proposed fusion method can incrementally find the redundant Gaussians using efficient pixel-wise alignment, and perform point-level feature fusion through a GRU network. 

\subsection{Weighted Floater Removal}

In feed-forward whole-scene reconstruction, the noise in the predicted depth maps is inevitable due to the complexity of scenarios (\textit{e.g.} Figure \ref{fig:floaters}), which results in severe floaters. Although the previous PTF module can mitigate 3D Gaussian redundancy and remove some floaters, it is sensitive to the floaters added later. Specifically, during the incremental process of PTF module, it can only remove floaters added before the current view. Conversely, in traditional TSDF Fusion method \cite{kinectfusion}, the multi-view depth maps are unprojected and averaged in the global voxel volumes before mesh extraction, thus storing the SDF values for the unoccupied voxels. However, adding extensive steps as TSDF Fusion to build voxel volumes and extract mesh can bring great time and computational cost, we thus propose an explicit approach to remove the floaters in the whole-scene reconstructed 3D Gaussian primitives, which can be seamlessly integrated with our method.

Specifically, after unprojecting and fusing whole scene input views using PTF module, we have the global 3DGS with centers $\bm{\mu}_g$, opacities $\bm{\alpha}_g$ and weights $\bm{\omega}_g$ of the whole scene. Then, we traverse the input views again to incrementally adjust the global Gaussian opaticies. 
In timestamp $t$, similarly to PTF, we project $\bm{\mu}_g$ to current view. For each pixel $i$ in the current view, similarly to Eq. \eqref{eq:projection}, we define $\bm{\mathcal{S}}_i'$ as the set of 3D Gaussian primitives that project onto pixel $i$.

To indicate Gaussian floaters, we search for pairs where the predicted depth value $d_l^t(j)$ is significantly larger than the rendered depth of nearest global Gaussian primitive. Such case is an indication that the corresponding global Gaussians may be floaters due to the noise in predicted depth maps. Given a pre-defined threshold $\delta$, potential floaters $m_i'$ in pixel $i$ are indicated as:
\begin{equation}
\label{alignment}
    m_i'=\begin{cases}
    \begin{array}{c@{\quad}l}
\mathop{\arg\min}\limits_{j\in\bm{\mathcal{S}}_i'}d^t_g(j), &\text{if}\;d_l^t(i)-\mathop{\min}\limits_{j\in\bm{\mathcal{S}}_i'}d^t_g(j)>\delta\\
\hfil \varnothing, &\text{otherwise}
\end{array}.
\end{cases}
\end{equation}

To eliminate such floaters without significantly affecting correctly localized Gaussian primitives, we introduce a weighted opacity reduction method based on accumulated weights:

\begin{equation}
\label{eq:wfr}
\bm{\alpha}_g(m_i')\leftarrow{}\frac{\bm{\tilde{\omega}}_g(m_i')\bm{\alpha}_g(m_i')}{\bm{\tilde{\omega}}_g(m_i')+\bm{\tilde{\omega}}_l(i)},
\end{equation}
where the weights are computed using a Neighbor Accumulation strategies to further leverage the Gaussian density information:

\begin{gather}
    \bm{\tilde{\omega}}_g(m_i')=\sum_{j\in\mathcal{J}(d_g^t(m_i'))}\bm{\omega}_g\left(j\right),\;\bm{\tilde{\omega}}_l(i)=\sum_{j\in\mathcal{J}(d_l^t(i))}\bm{\omega}_g\left(j\right),\\
    \text{where}\;\mathcal{J}(d)=\{j,|d - d_g^t(j)| < \delta\},
\end{gather}
such that we accumulate the weights of the neighboring gaussians of the global and local gaussians respectively, and decrease the opacities of the foreground gaussians accordingly. The insight of such neighbors accumulation is based on the observation that, the noisy predicted depth values are in minority, thus most 3D Gaussians should be located near the true surface. Therefore, accumulating the weights of neighboring Gaussians is to adaptively assign higher weights to the correct 3D Gaussians in Eq. \ref{eq:wfr}. 
Intuitively speaking, $\bm{\tilde{\omega}}_g(m_i')$ measures the accumulated weights of global Gaussian primitives in the neighbor of indicated floater $m_i'$. $\bm{\tilde{\omega}}_g(m_i')$ increases when more Gaussian primitives with higher weights distribute around $m_i'$, which means $m_i'$ is more reliable and will increase corresponding opacity $\bm{\alpha}_g(m_i')$.
In contrast, $\bm{\tilde{\omega}}_l(i)$ reflects the reliability of the predicted depth $d_l^t(i)$ also by accumulating the weights of nearby global Gaussian primitives. Since the floater $m_i'$ is indicated by differences between predicted depth and Gaussian primitive depth according to Eq.~\eqref{alignment}, higher reliability of predicted depth means the global Gaussian is more likely a floater. Therefore, as shown in Eq.~\eqref{eq:wfr}, higher $\bm{\tilde{\omega}}_l(i)$ leads to lower opacity. 

Overall, the Weighted Floater Removal strategy can be regarded as an explicit version of TSDF Fusion, which can effectively reduce the influence from floaters through checking multi-view depth consistency. Furthermore, it avoids leaving holes in the renderings thanks to its weight-based opacity decrease, and can be seamlessly integrated into our feed-forward 3DGS framework.

\subsection{Training}
\noindent\textbf{Loss Functions.} After predicting the 3D Gaussian primitives, we render from novel views following the rendering equations in Eq. \eqref{eq:render}. We train our framework using photometric losses, \textit{i.e.} a combination of MSE loss and LPIPS \cite{lpips} loss, with weights of 1 and 0.05 following \cite{pixelsplat, mvsplat}. Since our model can operate in feed-forward scheme, we can optionally leverage the ground truth depth images to directly regularize the $\bm{D}^t$ from Eq. \eqref{eq:depth}, which is discussed in \ref{section:depth}. Note that although we use lower resolution map for unprojection (\textit{cf}. \ref{decoding}), we still supervise the output on full resolution to encourage high-resolution renderings. 

\vspace{3pt}\noindent\textbf{Free-View Training.} 
We propose a Free-View Training (FVT) strategy to add more 
geometrical constraints on the localization of 3D Gaussian primitives, and to disentangle the performance of generalizable 3DGS with specific number of input views. To this end, we randomly sample $T$ number of context views (in experiments we set $T$ between $2$ and $8$), and supervise the image renderings in the broader view interpolations.
The long sequence training is made feasible due to our efficient feature extraction and aggregation. We empirically find that FVT significantly contributes to depth estimation and adapting to longer input sequences (\textit{cf.} Table \ref{tab:depth}, \ref{tab:ablation}). 

\vspace{3pt}\noindent\textbf{Depth-Regularized Fine-Tuning.} 
In whole-scene feed-forward reconstruction, some repeatedly observed regions can instead become blurry and unclear due to the accumulation of reconstruction errors.
Therefore, we optionally integrate a depth-regularized fine-tuning step to further refine the 3DGS primitives while preserving accurate 3D geometry. Leveraging the predicted 3D Gaussians as initialization to render depth maps at all training views $\left\{\bm{\mathcal{D}}_r^n\right\}_{n=1}^N$, we can calculate the depth regularization:
\begin{equation}
    \mathcal{L}_{depth}^{ft}=\frac{1}{N}\sum_{n=1}^N\Vert \hat{\bm{\mathcal{D}}}_r^n-\bm{\mathcal{D}}_r^n\Vert_1.
\end{equation}

\noindent The final regularized fine-tuning loss $\mathcal{L}_{rft}$ would be:
\begin{equation}
    \mathcal{L}_{rft} = (1-\lambda_1)\mathcal{L}_{color}^{ft}+\lambda_1\mathcal{L}_{ssim}^{ft}+\lambda_2\mathcal{L}_{depth}^{ft},
\end{equation}
where we set $\lambda_1$ and $\lambda_2$ as $0.2$ and $0.1$, respectively. The $\left\{\bm{\mathcal{D}}_r^n\right\}_{n=1}^N$ are dense and multi-view consistent, providing powerful depth priors to the per-scene fine-tuning while avoiding harming the rendering quality (\textit{c.f.} Table \ref{tab:finetune}, \ref{tab:whole_scannetpp}). 

\section{Experiments}
\begin{table*}[t]
    \small
    \centering
    \captionsetup{font=small}
    \caption{\small \textbf{Few-Views (2, 3 views) Novel View Interpolation results on ScanNet \cite{dai2017scannet}.} FreeSplat-\textit{fv} is trained with our FVT strategy, and the other methods are all trained on specific number of views to form a complete comparison. Time(s) indicates the total time of encoding input images and rendering one image. Within each column, \colorbox{red!30}{best}, \colorbox{orange!30}{runner-up}, and \colorbox{yellow!35}{second runner-up} results are differently colored.}
    \setlength{\tabcolsep}{1.6mm}{
    \begin{tabular}{c|cccccc|cccccc}
        \toprule

         \multirow{2}{*}{Method}&\multicolumn{6}{c|}{2 views} & \multicolumn{6}{c}{3 views}\\
         & PSNR$\uparrow$& SSIM$\uparrow$& LPIPS$\downarrow$& Time(s)$\downarrow$ & \#GS(K) & FPS$\uparrow$ & PSNR$\uparrow$& SSIM$\uparrow$& LPIPS$\downarrow$& Time(s)$\downarrow$ & \#GS(K) & FPS$\uparrow$ \\
         \midrule
         NeuRay \cite{neuray}& 25.65 & \cellcolor{red!30}0.840 & 0.264  & 3.103 & -& 0.3 & 25.47&\cellcolor{red!30}0.843&0.264&4.278&-& 0.2\\
         \midrule
         pixelSplat \cite{pixelsplat} & 26.03 & 0.784 & 0.265   & 0.289 & 1180 &128& 25.76 & 0.782 & 0.270 & 0.272 & 1769 & 100\\
         MVSplat \cite{mvsplat} & 27.27 & 0.822 & 0.221   & 0.117& 393 & 344 & 26.68 & 0.814 &0.235 & 0.192 & 590 & 269\\
         PixelGaussian \cite{pixelgaussian} & 26.63 & 0.805 & 0.260   & 0.222& 672 & 234 & 25.74 & 0.788 &0.291 & 0.354 & 1156 & 221\\
         \midrule
         \textbf{FreeSplat-\textit{spec}} & \cellcolor{orange!30}28.08 & \cellcolor{yellow!35}0.837 & \cellcolor{red!30}0.211  & \cellcolor{orange!30}0.103 &  278 & \cellcolor{yellow!35}496& \cellcolor{red!30}27.45 & \cellcolor{orange!30}0.829 & \cellcolor{red!30}0.222 & \cellcolor{orange!30}0.121 & 382 & \cellcolor{orange!30}507\\
         \textbf{FreeSplat-\textit{fv}} & \cellcolor{yellow!35}27.67 & 0.830& \cellcolor{yellow!35}0.215& \cellcolor{yellow!35}0.104& 279 & \cellcolor{orange!30}502 & \cellcolor{yellow!35}27.34& 0.826 & \cellcolor{yellow!35}0.226 & \cellcolor{yellow!35}0.122 & 390 & \cellcolor{yellow!35}506\\
         \textbf{FreeSplat++} & \cellcolor{red!30}28.19 & \cellcolor{orange!30}0.838&\cellcolor{orange!30}0.212 & \cellcolor{red!30}0.096&68&\cellcolor{red!30}556  &\cellcolor{red!30}27.45 & \cellcolor{orange!30}0.829 & \cellcolor{orange!30}0.223 & \cellcolor{red!30}0.113 &89 & \cellcolor{red!30}554\\
         \bottomrule
    \end{tabular}
    }
    \label{tab:inter_scannet}
\end{table*}

\subsection{Experimental Settings}
\noindent\textbf{Datasets.} We use two real-world indoor dataset ScanNet \cite{dai2017scannet} and ScanNet++ \cite{scannet++} for training. ScanNet is a large RGB-D dataset containing $1,513$ indoor scenes with camera poses, and we follow \cite{nerfusion, gao2023surfelnerf} to use 100 scenes for training and 8 scenes for testing. ScanNet++ is a high-quality indoor dataset with over 450 indoor scenes. We use the official training split as the training set and select 4 scenes from the official validation split to evaluate whole scene reconstruction. To evaluate the generalization ability of our model, we further perform zero-shot evaluation on the synthetic indoor dataset Replica \cite{replica}, for which we follow \cite{semantic_nerf} to select 8 scenes for testing.

\label{sec:implementation}
\noindent\textbf{Implementation Details.} Our model is trained end-to-end using Adam \cite{adam} optimizer with an initial learning rate of $1e-4$ and cosine decay following \cite{mvsplat}. To compare with baselines \cite{pixelsplat, mvsplat} on higher-resolution inputs while avoiding GPU OOM, all input images are resized to $384\times 512$ and batch size is set to 1. To evaluate the novel view synthesis performance given varying input sequence length, we set up various evaluation settings including few-view (2, 3 views) reconstruction, long sequence (10 views) reconstruction, and whole scene reconstruction. For long sequence reconstruction, we also select extrapolation views that are beyond the input view sequence. For whole scene reconstruction on ScanNet, we use the first 90\% of the whole sequence as input and sample views from the rest as extrapolation views. For per-scene 3DGS optimization, we follow the hyper-parameter setting as \cite{semanticgaussians}, and we follow \cite{nerfrpn} and \cite{roessle2022dense} for running COLMAP \cite{dai2017scannet} on ScanNet. For per-scene fine-tuning, we fine-tune our feed-forward 3DGS for $10,000$ iterations on ScanNet, and $15,000$ iterations on ScanNet++. In comparison, we run $30,000$ iterations for conventional per-scene optimized methods (3DGS \cite{3dgs}, VCR-GauS \cite{vcr}). Our experiments are conducted on single NVIDIA RTX A6000 GPU.


\vspace{3pt}\noindent\textbf{Baselines.} On feed-forward reconstruction setting, we compare with 3DGS-based methods pixelSplat \cite{pixelsplat}, MVSplat \cite{mvsplat}, PixelGaussian \cite{pixelgaussian}, and NeRF-based method NeuRay \cite{neuray}. For per-scene optimization results, we compare with the vanilla 3DGS \cite{3dgs}. We also compare with the models in our prior work FreeSplat \cite{freesplat}, including \textbf{FreeSplat-\textit{spec}} for vanilla FreeSplat training on specific numbers of views (2, 3 views), \textbf{FreeSplat-\textit{fv}} for vanilla FreeSplat with Free View Training. Note that our \textbf{FreeSplat++} is trained with Free View Training strategy to fit to arbitrary numbers of input views. For Per-Scene Fine-tuning, we use $_{ft}$ to represent vanilla 3DGS optimization, and $_{rft}$ to represent our regularized optimization. To evaluate image rendering quality, we report PSNR, SSIM \cite{ssim}, and LPIPS \cite{lpips}, and to evaluate geometric accuracy, we report rendered depth accuracy, \textit{i.e.} the Absolute Difference (Abs. Diff), Relative Difference (Rel. Diff), and threshold tolerance $\delta<1.25$, $\delta<1.1$.

\begin{table*}[t]
    \small
    \centering
    \captionsetup{font=small}
    \caption{\small \textbf{Long Sequence (10 views) Explicit Reconstruction results on ScanNet.} 
    }
    \setlength{\tabcolsep}{3.5mm}{
    \begin{tabular}{c|ccc|ccc|ccc}
        \toprule

         \multirow{2}{*}{Method}&\multirow{2}{*}{Time(s)$\downarrow$ }&\multirow{2}{*}{\#GS(K)}&\multirow{2}{*}{FPS$\uparrow$}&\multicolumn{3}{c}{View Interpolation} & \multicolumn{3}{c}{View Extrapolation}\\
         &  & & &PSNR$\uparrow$& SSIM$\uparrow$& LPIPS$\downarrow$& PSNR$\uparrow$& SSIM$\uparrow$& LPIPS$\downarrow$ \\
         \midrule
         pixelSplat \cite{pixelsplat} & 0.948& 5898& 55 & 21.26  & 0.714 & 0.396 & 20.70 & 0.687 & 0.429\\
         MVSplat \cite{mvsplat} & 1.178 & 1966 & 172 & 22.78  & 0.754 & 0.335 &21.60 & 0.729 & 0.365\\
         PixelGaussian \cite{pixelgaussian} & 2.442 & 2896 & 115 & 22.68  & 0.740 & 0.359 &21.36 & 0.710 & 0.388\\
         \midrule
         \textbf{FreeSplat-\textit{3views}} & \cellcolor{yellow!35}0.599 & 882 & \cellcolor{orange!30}342  &  \cellcolor{yellow!35}25.15 & \cellcolor{yellow!35}0.800 & \cellcolor{yellow!35}0.278 & \cellcolor{yellow!35}23.78 & \cellcolor{yellow!35}0.774 & \cellcolor{yellow!35}0.309\\
         \textbf{FreeSplat-\textit{fv}} & \cellcolor{orange!30}0.596 & 899 & \cellcolor{yellow!35}338 &  \cellcolor{orange!30}25.90 & \cellcolor{orange!30}0.808 & \cellcolor{orange!30}0.252 & \cellcolor{orange!30}24.64 & \cellcolor{orange!30}0.786 & \cellcolor{orange!30}0.277\\
         \textbf{FreeSplat++} & \cellcolor{red!30}0.586 & 185 & \cellcolor{red!30}547 &  \cellcolor{red!30}26.15 & \cellcolor{red!30}0.812 & \cellcolor{red!30}0.249 & \cellcolor{red!30}24.83 & \cellcolor{red!30}0.789 & \cellcolor{red!30}0.274\\
         \bottomrule
    \end{tabular}
    }
    \label{tab:long}
\end{table*}

\begin{table*}[h]
    \small
    \centering
    \captionsetup{font=small}
    \caption{\small \textbf{Novel View Depth Rendering results on ScanNet.}
    }
    \setlength{\tabcolsep}{1.9mm}{
    \begin{tabular}{c|ccc|ccc|ccc}
        \toprule

         \multirow{2}{*}{Method}&\multicolumn{3}{c|}{2 views} & \multicolumn{3}{c|}{3 views}& \multicolumn{3}{c}{10 views$^\dag$}\\
         &Abs Diff$\downarrow$&Abs Rel$\downarrow$& $\delta<1.25\uparrow$&Abs Diff$\downarrow$&Abs Rel$\downarrow$& $\delta<1.25\uparrow$&Abs Diff$\downarrow$&Abs Rel$\downarrow$&$\delta<1.25\uparrow$\\
         \midrule
         NeuRay \cite{neuray}& 0.358 & 0.200 & 0.755  & 0.231 & 0.117 & 0.873 & 0.202 & 0.108 & 0.875\\
         \midrule
         pixelSplat \cite{pixelsplat} & 1.205 & 0.745 & 0.472  & 0.698 & 0.479 &0.836 & 0.970 & 0.621 & 0.647\\
         MVSplat \cite{mvsplat} &0.192 & 0.106 & 0.912  &0.164 &0.079 & \cellcolor{yellow!35}0.929 & 0.142 & 0.080 & 0.914\\
         PixelGaussian \cite{pixelgaussian} &  0.176 &0.097 & 0.893 & 0.166 & 0.080 & 0.922 & 0.172 & 0.098 & 0.866\\
         \midrule
         \textbf{FreeSplat-\textit{spec}} & \cellcolor{yellow!35}0.157 & \cellcolor{yellow!35}0.086 & \cellcolor{yellow!35}0.919  & \cellcolor{orange!30}0.161 & \cellcolor{orange!30}0.077 & \cellcolor{orange!30}0.930 & \cellcolor{yellow!35}0.120 & \cellcolor{yellow!35}0.070 & \cellcolor{yellow!35}0.945\\
         \textbf{FreeSplat-\textit{fv}} & \cellcolor{red!30}0.153 & \cellcolor{red!30}0.085 & \cellcolor{red!30}0.923  & \cellcolor{yellow!35}0.162 & \cellcolor{orange!30}0.077 &0.928 & \cellcolor{orange!30}0.097 & \cellcolor{orange!30}0.059 & \cellcolor{orange!30}0.961\\
         \textbf{FreeSplat++} & \cellcolor{red!30}0.153 & \cellcolor{red!30}0.085 & \cellcolor{orange!30}0.920 & \cellcolor{red!30}0.154 & \cellcolor{red!30}0.073 & \cellcolor{red!30}0.933 &\cellcolor{red!30}0.088 & \cellcolor{red!30}0.054 & \cellcolor{red!30}0.968\\
         \bottomrule
    \end{tabular}
    }
    \label{tab:depth}
\end{table*}
\begin{table*}[t]
    \small
    \captionsetup{font=small}
    \caption{\small \textbf{Zero-Shot Transfer Results on Replica \cite{replica}.} }
    \centering
    \setlength{\tabcolsep}{1.2mm}{
    \begin{tabular}{c|cccccc|cccccc}
        \toprule
         \multirow{2}{*}{Method}&\multicolumn{6}{c|}{3 Views} & \multicolumn{6}{c}{10 Views}\\
         & PSNR$\uparrow$& SSIM$\uparrow$& LPIPS$\downarrow$&$\delta<1.25\uparrow$& \#GS(K) & FPS$\uparrow$ & PSNR$\uparrow$& SSIM$\uparrow$& LPIPS$\downarrow$&$\delta<1.25\uparrow$& \#GS(K)& FPS$\uparrow$ \\
         \midrule
         pixelSplat \cite{pixelsplat} & 26.24 & 0.829  & 0.229 & 0.576 & 1769 &112&19.23&0.719&0.414&0.375&5898 & 43\\
         MVSplat \cite{mvsplat} & 26.16 & 0.840 & \cellcolor{orange!30}0.173 & 0.670 & 590 &253 & 18.66 & 0.717 &0.360 & 0.565 & 1966 & 142\\
         PixelGaussian \cite{pixelgaussian} & 23.38 & 0.795 & 0.264 & 0.586 & 1171 & 164 & 16.61 & 0.667 & 0.434 &0.328& 2952 & 94\\
         \midrule
         \textbf{FreeSplat-\textit{spec}} & \cellcolor{red!30}26.98 & \cellcolor{red!30}0.848 & \cellcolor{red!30}0.171 & \cellcolor{orange!30}0.682 & 423 & \cellcolor{orange!30}375 &\cellcolor{yellow!35}21.11 & \cellcolor{yellow!35}0.762 & \cellcolor{yellow!35}0.312 & \cellcolor{yellow!35}0.720 & 1342 & \cellcolor{yellow!35}276 \\
         \textbf{FreeSplat-\textit{fv}} & \cellcolor{yellow!35}26.64 & \cellcolor{orange!30}0.843 & \cellcolor{yellow!35}0.184 & \cellcolor{orange!30}0.682 & 421 & \cellcolor{yellow!35}369 & \cellcolor{orange!30}21.95 & \cellcolor{orange!30}0.777 & \cellcolor{red!30}0.290 & \cellcolor{orange!30}0.742 & 1346 & \cellcolor{orange!30}280\\
         \textbf{FreeSplat++} & \cellcolor{orange!30}26.75 & \cellcolor{orange!30}0.843 & 0.190 & \cellcolor{red!30}0.695 & 90 & \cellcolor{red!30}463 & \cellcolor{red!30}22.36 & \cellcolor{red!30}0.780 & \cellcolor{orange!30}0.292 & \cellcolor{red!30}0.749 & 224 & \cellcolor{red!30}440 \\
         \bottomrule
    \end{tabular}
    }
    \label{tab:replica}
\end{table*}
\subsection{Region Reconstruction Results}
\noindent\textbf{Few-View Reconstruction Results.} 
The few-view reconstruction results are shown in Table \ref{tab:inter_scannet}. FreeSplat-\textit{spec}, which is trained on specific numbers of views (2, 3), achieves improved rendering quality over the baselines, with higher inference and rendering efficiency. FreeSplat-\textit{fv} can effectively fit to 2 and 3 input views scenarios and produce comparable performance as FreeSplat-\textit{spec}. Furthermore, our FreeSplat++ offers competitive rendering quality with fewer number of gaussians, leading to further improved efficiency.

\vspace{3pt}\noindent\textbf{Long Sequence Results.} 
The long sequence reconstruction results are shown in Table \ref{tab:long}. To form a fair comparison, we compare both FreeSplat-\textit{3views} and FreeSplat-\textit{fv} results with the baselines. FreeSplat-\textit{3views} can significantly outperform the others in both rendering quality and efficiency. It reveals that our method can effectively takes in longer sequence inputs thanks to our adaptive cost volume formulation and Pixel-wise Triplet Fusion strategy. On the other hand, trained on longer sequence reconstruction thanks to our low-cost backbone, FreeSplat-\textit{fv} consistently outperforms our 3-views version. Our PTF module can also reduce the number of Gaussians by around $55.0\%$, which becomes indispensable in long sequence reconstruction due to the pixel-wise unprojection nature of generalizable 3DGS. Furthermore, our FreeSplat++ achieves the state-of-the-art rendering quality with only $20.6\%$ number of gaussians as FreeSplat-\textit{3views}.

\vspace{3pt}\noindent\textbf{Novel View Depth Estimation Results.} We also investigated the accuracy of 3D Gaussian localization of different methods through comparing their depth rendering accuracy. As shown in Table \ref{tab:depth}, we find that FreeSplat consistently outperforms pixelSplat and MVSplat in predicting accurately localized 3D Gaussians. Furthermore, FreeSplat++ reaches $96.8\%$ of $\delta<1.25$, enabling accurate unsupervised depth estimation which is comparable to supervised methods \cite{deepvideomvs, simplerecon}. 
The results also indicate that our FreeSplat++ can more effectively integrate longer sequence inputs for accurate depth estimation, which serves as a basis for whole scene reconstruction.

\begin{table*}[t!]
    \small
    \centering
    \captionsetup{font=small}
    \caption{\small \textbf{Whole Scene Reconstruction results on ScanNet.} iPSNR and ePSNR respectively represent the PSNR on the interpolated and extrapolated views. pixelSplat* uses deterministic mode, \textit{i.e.} predicting one Gaussian per pixel, due to memory limit.}
    \setlength{\tabcolsep}{3.3mm}{
    \begin{tabular}{c|c|ccccc|ccc}
        \toprule

         &Method&iPSNR$\uparrow$ & ePSNR$\uparrow$ & SSIM$\uparrow$ & LPIPS$\downarrow$& $\delta<1.1\uparrow$&Time(s)$\downarrow$&\#GS(M)&FPS$\uparrow$\\
         \midrule
         \parbox[t]{2mm}{\multirow{6}{*}{\rotatebox[origin=c]{90}{Generalizable}}}
         &pixelSplat* \cite{pixelsplat} & 15.54 & 13.47 & 0.557 & 0.608 & 0.023 & 27.5 & 29.88 & 25 \\
         &MVSplat \cite{mvsplat} & 16.51 & 13.67 & 0.591 & 0.541 & 0.323 & 39.5 & 29.88 & 25 \\
         &PixelGaussian \cite{pixelgaussian} & 16.33 & 13.40 & 0.601 & 0.549 & 0.282 & 47.9 & 45.83 & 16 \\
         &\textbf{FreeSplat-\textit{3views}} & \cellcolor{yellow!35}20.35 & \cellcolor{yellow!35}16.67 & \cellcolor{yellow!35}0.706 & \cellcolor{yellow!35}0.423 & \cellcolor{yellow!35}0.573 & \cellcolor{yellow!35}24.4 & 11.20 & \cellcolor{orange!30}66\\
         &\textbf{FreeSplat-\textit{fv}} & \cellcolor{orange!30}21.80 & \cellcolor{orange!30}18.09 & \cellcolor{orange!30}0.739 & \cellcolor{orange!30}0.360 & \cellcolor{orange!30}0.771 & \cellcolor{orange!30}23.3 & 11.27 & \cellcolor{yellow!35}62 \\
         &\textbf{FreeSplat++} & \cellcolor{red!30}23.29 & \cellcolor{red!30}19.44& \cellcolor{red!30}0.771 & \cellcolor{red!30}0.320 & \cellcolor{red!30}0.904 & \cellcolor{red!30}20.9 & 1.46 & \cellcolor{red!30}453 \\
         \midrule
         \midrule
         \parbox[t]{2mm}{\multirow{5.5}{*}{\rotatebox[origin=c]{90}{Per-Scene}}}&3DGS \cite{3dgs} & \cellcolor{yellow!35}26.66 & \cellcolor{yellow!35}19.45 & \cellcolor{yellow!35}0.835 & \cellcolor{yellow!35}0.328 & 0.417  & 3747& 1.59 & 409\\
         & VCR-GauS \cite{vcr} & 24.14 & 15.33 & 0.791 & 0.393 & \cellcolor{yellow!35}0.496 & 3803 & 1.92 & \cellcolor{orange!30}412 \\
         \cmidrule{2-10}
         &MVSplat$_{rft}$ \cite{mvsplat}&26.05&18.33&0.820&0.341&0.468&\cellcolor{yellow!35}1454&23.05&54\\
         &\textbf{FreeSplat-\textit{fv}}$_{rft}$& \cellcolor{orange!30}26.90 & \cellcolor{orange!30}20.05 & \cellcolor{orange!30}0.836 & \cellcolor{red!30}0.310 & \cellcolor{orange!30}0.780 & \cellcolor{orange!30}1390 & 8.91 &\cellcolor{yellow!35}110\\
         &\textbf{FreeSplat++}$_{rft}$ & \cellcolor{red!30}26.92 & \cellcolor{red!30}20.36 & \cellcolor{red!30}0.837 & \cellcolor{orange!30}0.316 & \cellcolor{red!30}0.909 & \cellcolor{red!30}616 & 1.26 & \cellcolor{red!30}508 \\
         \bottomrule
    \end{tabular}
    }
    \label{tab:whole_scannet}
\end{table*}

\begin{table*}[t!]
    \small
    \centering
    \captionsetup{font=small}
    \caption{\small \textbf{Whole Scene Reconstruction results on ScanNet++ \cite{scannet++}.}}
    \setlength{\tabcolsep}{3.3mm}{
    \begin{tabular}{c|c|ccccc|ccc}
        \toprule

         &Method&iPSNR$\uparrow$ & ePSNR$\uparrow$ & SSIM$\uparrow$ & LPIPS$\downarrow$& $\delta<1.1\uparrow$&Time(s)$\downarrow$&\#GS(M)&FPS$\uparrow$\\
         \midrule
         \parbox[t]{2mm}{\multirow{6}{*}{\rotatebox[origin=c]{90}{Generalizable}}}
         &pixelSplat* \cite{pixelsplat} & 10.70 & 10.37 & 0.497 & 0.663 & 0.000 & 59.0 & 54.46 & 8 \\
         &MVSplat \cite{mvsplat} & 11.10 & 10.62 & 0.497 & 0.648 & 0.028 & \cellcolor{orange!30}47.9 & 54.46 & 10  \\
         &PixelGaussian \cite{pixelgaussian} & 10.78 & 10.44 & 0.529 & 0.639 & 0.012 & 67.2 & 88.31 & 6  \\
         &\textbf{FreeSplat-\textit{3views}} & \cellcolor{yellow!35}16.03 & \cellcolor{yellow!35}14.00 & \cellcolor{yellow!35}0.652 & \cellcolor{yellow!35}0.470 & \cellcolor{yellow!35}0.490 & 75.8 & 16.04 & \cellcolor{orange!30}27 \\
         &\textbf{FreeSplat-\textit{fv}} & \cellcolor{orange!30}18.04 & \cellcolor{orange!30}15.90 & \cellcolor{orange!30}0.727 & \cellcolor{orange!30}0.370 & \cellcolor{orange!30}0.604 & \cellcolor{yellow!35}55.7 & 17.59 & \cellcolor{yellow!35}16\\
         &\textbf{FreeSplat++} & \cellcolor{red!30}22.63 & \cellcolor{red!30}19.51 & \cellcolor{red!30}0.829 & \cellcolor{red!30}0.261 & \cellcolor{red!30}0.890 & \cellcolor{red!30}42.3 & 1.76 & \cellcolor{red!30}227  \\
         \midrule
         \midrule
         \parbox[t]{2mm}{\multirow{5.5}{*}{\rotatebox[origin=c]{90}{Per-Scene}}}&3DGS \cite{3dgs}&\cellcolor{red!30}33.26 & \cellcolor{yellow!35}20.11 & \cellcolor{orange!30}0.963 & \cellcolor{orange!30}0.138 & 0.518 & \cellcolor{yellow!35}3007 & 0.48 & \cellcolor{red!30}365 \\
         & VCR-GauS \cite{vcr} & \cellcolor{yellow!35}32.43 & \cellcolor{orange!30}20.65 & \cellcolor{yellow!35}0.956 & \cellcolor{yellow!35}0.142 & \cellcolor{orange!30}0.777 & 3887 & 3.17 & \cellcolor{yellow!35}223\\
         \cmidrule{2-10}
         &MVSplat$_{rft}$ \cite{mvsplat} & 23.30 & 14.40 & 0.872 & 0.235 & 0.041 &3319 &18.77 &35 \\
         &\textbf{FreeSplat-\textit{fv}}$_{rft}$&30.85 & 19.17 & 0.939 & 0.157&\cellcolor{yellow!35}0.686&\cellcolor{orange!30}1624&3.53&162\\
         &\textbf{FreeSplat++}$_{rft}$ & \cellcolor{orange!30}33.05 & \cellcolor{red!30}21.35 & \cellcolor{red!30}0.966 & \cellcolor{red!30}0.109 & \cellcolor{red!30}0.908 & \cellcolor{red!30}949 & 1.51 & \cellcolor{orange!30}254  \\
         \bottomrule
    \end{tabular}
    }
    \label{tab:whole_scannetpp}
\end{table*}
\begin{table*}[t]
    \small
    \centering
    \captionsetup{font=small}
    \caption{\small \textbf{Ablation study of feed-forward modules with results of whole scene reconstruction.}}
    \setlength{\tabcolsep}{0.4mm}{
    \begin{tabular}{c|cccccc|cccccc}
        \toprule
        \multirow{2}{*}{Method}&\multicolumn{6}{c|}{ScanNet}&\multicolumn{6}{c}{ScanNet++}\\
         &iPSNR$\uparrow$&ePSNR$\uparrow$& $\delta<1.1\uparrow$& Time(s)$\downarrow$&\#GS(M)&FPS$\uparrow$&iPSNR$\uparrow$&ePSNR$\uparrow$& $\delta<1.1\uparrow$& Time(s)$\downarrow$&\#GS(M)&FPS$\uparrow$\\
         \midrule
         \textit{wo}/ Cost Volume & 18.90 & 15.83 & 0.304 & \cellcolor{red!30}4.4 & 2.05 & 324 & 18.74 & 15.74 & 0.469 & \cellcolor{red!30}10.1 & 2.57 & 168 \\
         \textit{wo}/ Free View Training (8-views) & 22.90 & \cellcolor{yellow!30}19.32&  0.865 & 21.3 &1.54 & \cellcolor{yellow!30}408 & \cellcolor{yellow!30}22.35 & \cellcolor{yellow!30}19.22 & \cellcolor{yellow!30}0.878 & 44.5 & 1.83 & \cellcolor{yellow!30}204 \\
         \textit{wo}/ Fusion & 22.11 & 18.50 & 0.874 & 24.1 & 7.29 & 78 & 18.88 & 16.63 & 0.712 & 150.9 & 13.88 & 21 \\
         \textit{wo}/ Broader Fusion in Eq. \eqref{eq:alignment} & \cellcolor{yellow!30}22.94 & 19.11 & \cellcolor{yellow!30}0.890 & 21.2 & 2.59 & 214 & 20.74 & 17.90 & 0.842 & 45.3 & 4.37 & 78 \\
         \textit{wo}/ Floater Removal & 22.88 & 19.20 & 0.859 & \cellcolor{orange!30}19.6 & 1.46 & \cellcolor{orange!30}443&20.02&17.66 & 0.740 & \cellcolor{orange!30}41.5 & 1.76&\cellcolor{orange!30}220\\
         \textit{wo}/ Lower Resolution & \cellcolor{orange!30}23.21 & \cellcolor{red!30}19.48 & \cellcolor{red!30}0.907 & 24.5 & 6.27 & 120 & \cellcolor{red!30}22.70 & \cellcolor{red!30}19.57 & \cellcolor{red!30}0.907 & 133.8 & 7.21 & 62\\
         \midrule
         \textbf{FreeSplat++} & \cellcolor{red!30}23.29 & \cellcolor{orange!30}19.44 & \cellcolor{orange!30}0.904 &\cellcolor{yellow!30}20.9 & 1.46 & \cellcolor{red!30}453&\cellcolor{orange!30}22.63&\cellcolor{orange!30}19.51&\cellcolor{orange!30}0.890&\cellcolor{yellow!30}42.3&1.76&\cellcolor{red!30}227\\
         \bottomrule
    \end{tabular}
    }
    \label{tab:ablation}
\end{table*}

\vspace{3pt}\noindent\textbf{Zero-Shot Transfer Results on Replica.} We further evaluate the zero-shot transfer results through testing on Replica dataset using our models which are trained on ScanNet, as shown in Table \ref{tab:replica}. Our view interpolation and novel view depth estimation results still outperforms existing methods. The long sequence results degrade due to inaccurate depth estimation and domain gap, indicating potential future work in building more robust multi-view depth estimation in zero-shot transferring.

\subsection{Whole Scene Reconstruction Results}
To further evaluate the methods' ability to extend to extremely long input sequence, we conduct the challenging whole scene reconstruction experiments on ScanNet and ScanNet++, with results shown in Table \ref{tab:whole_scannet} and Table \ref{tab:whole_scannetpp}, respectively. In whole scene reconstruction, generalizable 3DGS methods face three major challenges: 1) Severe floaters due to noise in predicted depth maps; 2) Difficulty to apply full cross-attention between all input views due to its quadratically increasing GPU requirement; 3) Extremely large number of 3D Gaussians due to pixel-wise unprojection. On both datasets, the baseline methods all give significantly degraded performance due to severe floaters, and extremely large number of gaussians that consume enormous GPU memory to store and render. In comparison, FreeSplat-\textit{3views} and FreeSplat-\textit{fv} significantly outperforms the baselines iPSNR by over $4.02$ dB on ScanNet and over $4.93$dB on ScanNet++, which indicates that the framework design of our prior version can handle feed-forward whole scene reconstruction more much effectively. 

\begin{figure*}[t!]
    \centering
    \captionsetup{font=small}
    \includegraphics[width=1\textwidth]{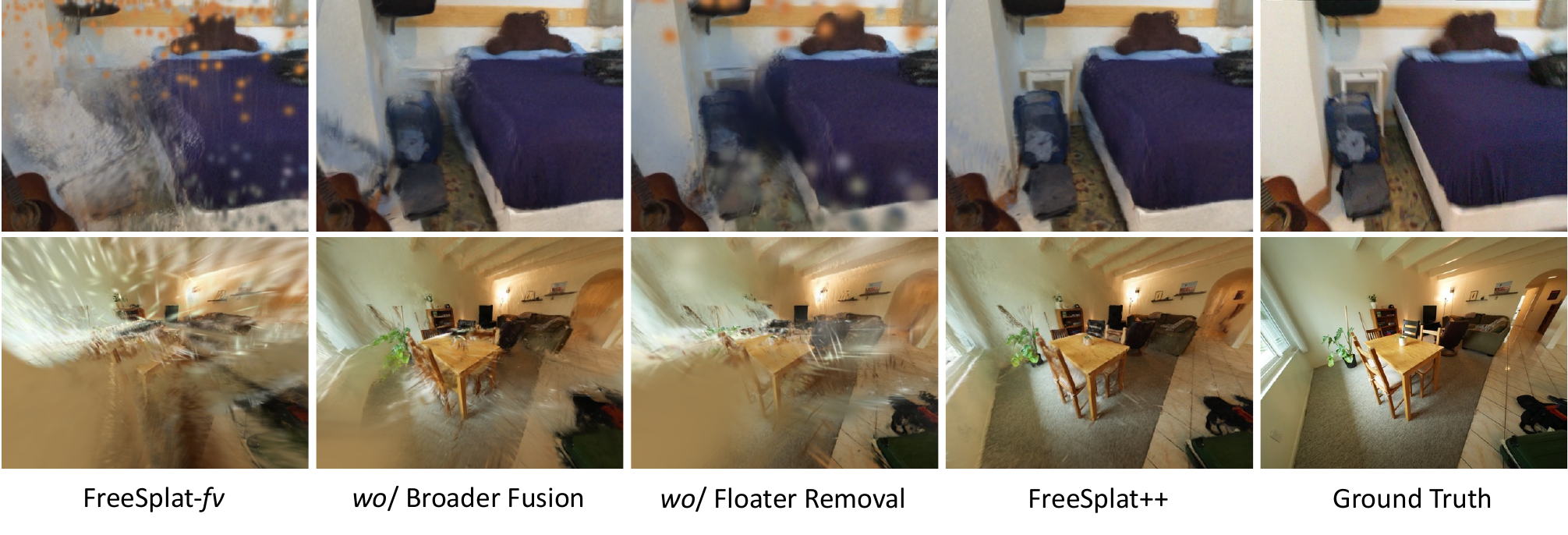}
    \caption{\textbf{Qualitative Ablation Study.} The first and second row are whole scene reconstruction results from ScanNet and ScanNet++, respectively.} 
    \label{fig:ablate}
\end{figure*}

Despite these advances, FreeSplat still struggles with unsatisfactory rendering quality due to floaters (cf. Figure \ref{fig:floaters}, \ref{fig:ablate}) and notably slow rendering speeds especially on ScanNet++. To this end, our FreeSplat++ can effectively improve the performance in such challenging scenario thanks to its carefully designed fusion and floater removal modules, leading to $1.49$ dB and $4.59$ dB iPSNR improvements on ScanNet and ScanNet++, respectively. Furthermore, the lower-resolution unprojection map further reduces the number of gaussians by $75\%$, which is a simple yet vital approach to keeping the number of gaussians affordable while not significantly affecting the performance. 

In terms of per-scene optimized results, after our depth-regularized fine-tuning process, our FreeSplat++$_{rft}$ gives similar or better interpolation performance than the vanilla 3DGS, improved extrapolation performance, and significantly enhanced geometry accuracy and training efficiency. On the other hand, comparing to vanilla 3DGS, MVSplat$_{rft}$ gives worse performance in both rendering quality and efficiency, and FreeSplat-\textit{fv}$_{rft}$ produces comparable or worse rendering quality, more accurate geometry, yet much slower rendering speed. Additionally, VCR-GauS \cite{vcr} gives worse color rendering quality and suboptimal depth accuracy, which is partially due to the multi-view inconsistency limitation of monocular-cue based per-scene optimization methods. 

\begin{figure*}[t!]
    \centering
    \includegraphics[width=1\textwidth]{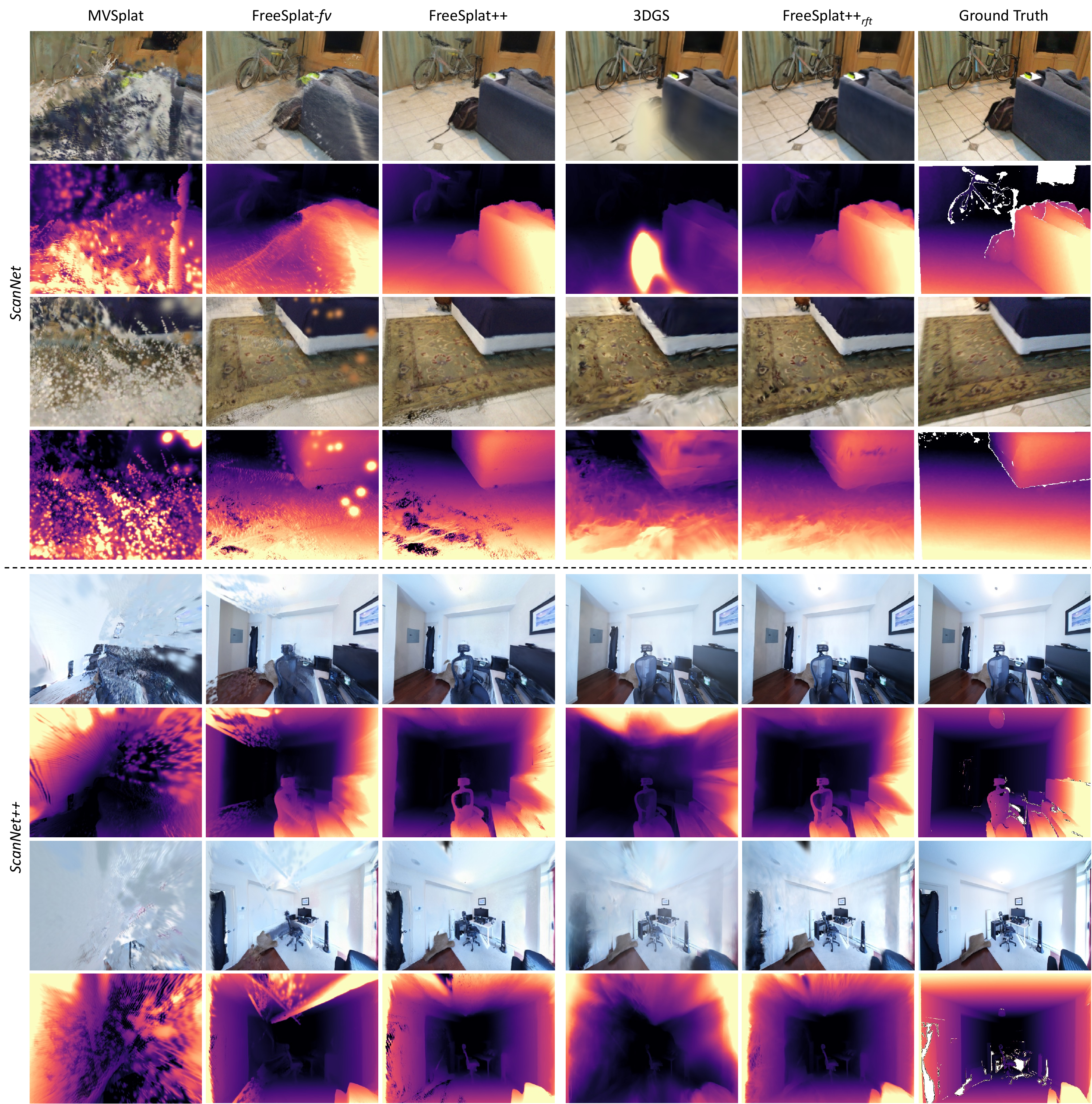}
    \caption{\textbf{Qualitative Results of Whole Scene Reconstruction on ScanNet \cite{dai2017scannet} and ScanNet++ \cite{scannet++}} For each dataset, the first two rows are view interpolation results (rendered color and depth images from novel views), and the last two rows are view extrapolation results.} 
    \label{fig:whole_qual}
\end{figure*}

\begin{figure*}[t!]
    \centering
    \captionsetup{font=small}
    \includegraphics[width=1\textwidth]{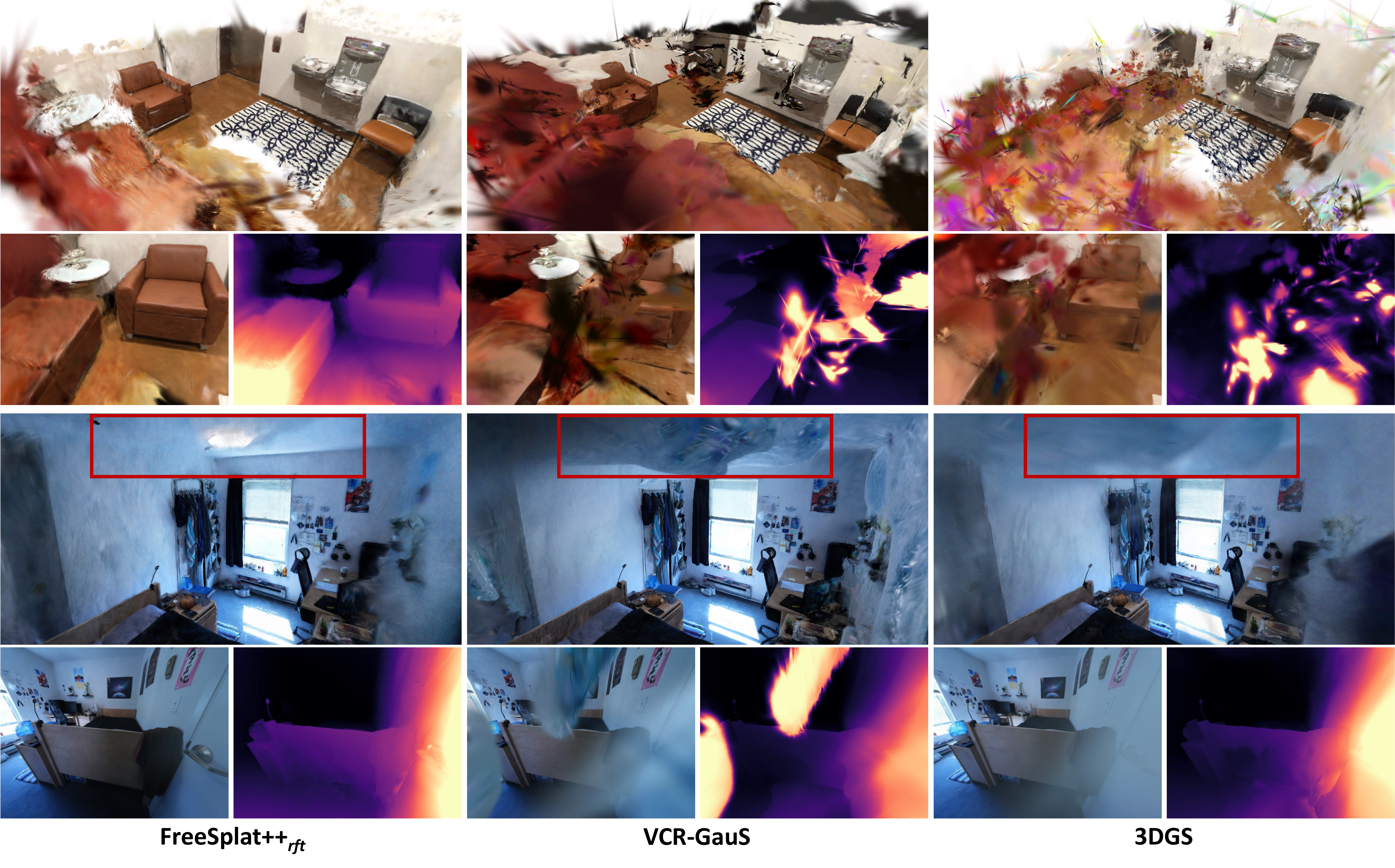}
    \caption{\textbf{Visualization of whole-scene 3D Gaussians and renderings.} Conventional per-scene optimized methods can suffer from inaccurate COLMAP point cloud initialization. Additionally, 3DGS lacks depth regularization and struggles in view extrapolation, while monocular-cue based method like VCR-GauS \cite{vcr} offers suboptimal performance due to the inconsistent monocular priors.} 
    \label{fig:whole_gs}
\end{figure*}

These findings underscore the indispensable role of our improved framework in facilitating both generalizable 3DGS prediction and per-scene fine-tuning. The qualitative results, showcased in Figure \ref{fig:whole_qual} and Figure \ref{fig:whole_gs}, highlight FreeSplat++'s advances in floater removal and precise scene reconstruction.



\begin{table*}[t!]
    \small
    \centering
    \captionsetup{font=small}
    \caption{\small \textbf{Ablation study of Per-Scene Optimization strategies.} Dense / Sparse represent fine-tuning using all the training views or only the input views to the feed-forward methods. The $N / M$ iterations represent training N iterations for feed-forward methods and M iterations for vanilla 3DGS, respectively.}
    \setlength{\tabcolsep}{1.1mm}{
    \begin{tabular}{c|cccccc|cccccc}
        \toprule
         \multirow{2}{*}{Method}&\multicolumn{6}{c|}{Dense ($10K / 30K$ iterations)} & \multicolumn{6}{c}{Sparse ($2K / 30K$ iterations)}\\
         & iPSNR$\uparrow$& ePSNR$\uparrow$& LPIPS$\downarrow$&$\delta<1.1\uparrow$& Time(s)$\downarrow$& \#GS(M) & iPSNR$\uparrow$& ePSNR$\uparrow$& LPIPS$\downarrow$&$\delta<1.1\uparrow$& Time(s)$\downarrow$& \#GS(M)\\
         \midrule
         MVSplat$_{ft}$ \cite{mvsplat}&26.76&19.82&\cellcolor{yellow!35}0.321&0.595&934&4.15&24.98&19.62&0.328&0.667&279&22.02\\
         \textbf{FreeSplat-\textit{fv}}$_{ft}$&\cellcolor{yellow!35}26.90&\cellcolor{yellow!35}20.15& \cellcolor{orange!30}0.320&\cellcolor{yellow!35}0.768&\cellcolor{yellow!35}830&2.77&\cellcolor{yellow!35}25.36&\cellcolor{yellow!35}19.67&\cellcolor{orange!30}0.317&\cellcolor{yellow!35}0.795&\cellcolor{yellow!35}249&8.75\\
         \textbf{FreeSplat++}$_{ft}$ & \cellcolor{red!30}26.94 & \cellcolor{orange!30}20.25 & 0.323 & \cellcolor{orange!30}0.842 &\cellcolor{red!30}593& 1.26&\cellcolor{orange!30}25.55&\cellcolor{orange!30}20.01&\cellcolor{yellow!35}0.319&\cellcolor{orange!30}0.878&\cellcolor{orange!30}82&1.60\\
         \textbf{FreeSplat++}$_{rft}$ & \cellcolor{orange!30}26.92 &  \cellcolor{red!30}20.36 & \cellcolor{red!30}0.316 & \cellcolor{red!30}0.909 &\cellcolor{orange!30}616& 1.26&\cellcolor{red!30}25.60&\cellcolor{red!30}20.03&\cellcolor{red!30}0.315&\cellcolor{red!30}0.899&\cellcolor{red!30}69&1.35 \\
         \midrule
         3DGS \cite{3dgs}& 26.66&19.45&0.328&0.417&3747&1.59&23.78&19.12&0.372&0.345&3634&1.39\\
         \bottomrule
    \end{tabular}
    }
    \label{tab:finetune}
\end{table*}


\subsection{Ablation Study}
\noindent\textbf{Framework Design.} An ablation study was conducted to validate the design of our framework, as illustrated in Table \ref{tab:ablation} and Figure \ref{fig:ablate}. First of all, the results highlight the the vital role of the cost volume in depth estimation. Secondly, our Free-View Training strategy can improve the performance over simply training on 8 views, affirming the efficacy of fitting the model to arbitrary numbers of views. On the other hand, the fusion module contributes significantly in both rendering quality and efficiency, since it can merge overlapping gaussian latent features, remove foreground floaters and accumulate gaussian weights to assist the subsequent floater removal process. In terms of our newly improved modules in FreeSplat++, broader fusion range and the floater removal module can work synergistically to significantly remove cross-view floaters to improve geometry accuracy. Additionally, the adoption of a lower-resolution map, while resulting in a moderate performance trade-off, substantially boosts overall efficiency. Furthermore, comparing the inference time of FreeSplat++ and \textit{wo}/ Floater Removal indicates that the average time cost of the floater removal module is around 1 second per scene, demonstrating its high efficiency.

\begin{table}[t]
    \small
    \centering
    \captionsetup{font=small}
    \caption{\small \textbf{Ablation study of Floater Removal strategies.} $^\dag$ directly uses $\bm{\omega}^t_g$ and $\bm{\omega}^t_l$ in Eq. \ref{eq:wfr}.}
    \setlength{\tabcolsep}{0.5mm}{
    \begin{tabular}{c|cc|cc}
        \toprule

         \multirow{2}{*}{Method}&\multicolumn{2}{c|}{ScanNet}&\multicolumn{2}{c}{ScanNet++}\\
         &iPSNR$\uparrow$ & $\delta<1.1\uparrow$&iPSNR$\uparrow$ & $\delta<1.1\uparrow$\\
         \midrule
         \textit{wo}/ Floater Removal& 22.88 & 0.859 & 20.02 & 0.740\\
         \midrule
         Direct Removal ($\delta=0.3$) & 22.78 & \cellcolor{yellow!30}0.893 & 22.20 & \cellcolor{yellow!30}0.853 \\
         Direct Removal ($\delta=0.1$) & 20.97 & 0.873 & 21.52 & 0.788 \\
         Uniform Weights & \cellcolor{yellow!30}22.94 & 0.892& \cellcolor{yellow!30}22.35 & 0.846  \\
         \textit{wo}/ Neighbors Accumulate$^\dag$ & \cellcolor{orange!30}23.21 & \cellcolor{orange!30}0.899 & \cellcolor{orange!30}22.38 & \cellcolor{orange!30}0.886 \\
         \midrule
         \textbf{FreeSplat++} & \cellcolor{red!30}23.29 & \cellcolor{red!30}0.904 & \cellcolor{red!30}22.63 & \cellcolor{red!30}0.890 \\
         \bottomrule
    \end{tabular}
    }
    \label{tab:wfr}
\end{table}

\vspace{3pt}\noindent\textbf{Floater Removal Strategies.} We also evaluate different floater removal strategies as shown in Table \ref{tab:wfr}. The direct removal strategy can improve the geometry accuracy, while harming the rendering quality due to its brute-force way of removing floaters that leaves holes in the rendered images. On the other hand, setting uniform weights in Eq. \eqref{eq:wfr} still struggles to improve the rendering quality. Furthermore, the neighbors accumulation strategy leverages the neighboring information to adaptively assign weights in Eq. \eqref{eq:wfr}, further improving the performance.

\vspace{3pt}\noindent\textbf{Per-Scene Finetuning Strategies.} We further explore different per-scene fine-tuning strategies as shown in Table \ref{tab:finetune}, in which we evaluate both dense and sparse per-scene optimization. The results indicate that our depth-regularized fine-tuning process is important in maintaining the accurate geometry initialization, and baseline methods still perform considerable worse than ours after vanilla 3DGS fine-tuning, especially on efficiency and geometric accuracy. In addition, the sparse fine-tuning results demonstrate the more important role of priors when given partial training data.

\begin{table}[t]
    \small
    \centering
    \captionsetup{font=small}
    \caption{\small \textbf{Ablation study of optional depth supervision during feed-forward framework training.} $\delta_p$ and $\delta_r$ indicate depth accuracy at predicted input views and rendered target views, respectively.}
    \setlength{\tabcolsep}{0.2mm}{
    \begin{tabular}{c|ccc|ccc}
        \toprule

         \multirow{2}{*}{Depth Sup.}&\multicolumn{3}{c|}{ScanNet}&\multicolumn{3}{c}{ScanNet++}\\
         &\footnotesize{iPSNR$\uparrow$} &\footnotesize{$\delta_p<1.1\uparrow$} & \footnotesize{$\delta_r<1.1\uparrow$}&\footnotesize{iPSNR$\uparrow$}&\footnotesize{$\delta_p<1.1\uparrow$}  & \footnotesize{$\delta_r<1.1\uparrow$}\\
         \midrule
         \XSolidBrush & \cellcolor{red!30}23.29 & \cellcolor{orange!30}0.875&\cellcolor{red!30}0.904 & \cellcolor{orange!30}22.63 & \cellcolor{orange!30}0.861 &\cellcolor{orange!30}0.890 \\
         \Checkmark & \cellcolor{orange!30}23.10 & \cellcolor{red!30}0.887&\cellcolor{orange!30}0.899 & \cellcolor{red!30}22.71 & \cellcolor{red!30}0.944&\cellcolor{red!30}0.942 \\
         \bottomrule
    \end{tabular}
    }
    \label{tab:depthsup}
\end{table}

\vspace{3pt}\noindent\textbf{Optional Depth Supervision.}
\label{section:depth}
As shown in Table \ref{tab:depthsup}, we compare the performance when adding depth supervision during feed-forward framework training. On ScanNet, the depth supervision leads to slight performance drop on iPSNR and $\delta_r<1.1$ while improving $\delta_p<1.1$. It indicates that depth supervision on ScanNet can slightly improve the predicted depth accuracy at input views, while the final reconstruction after fusion and floater removal is less accurate. The results also demonstrate the advantage of 3DGS-based unsupervised depth prediction. On the other hand, results on ScanNet++ indicate that the depth supervision can effectively improve the depth prediction accuracy, thus improving iPSNR and $\delta_r<1.1$ as well. These findings indicate that our feed-forward model can optionally utilize ground truth depth maps, when available, to improve reconstruction accuracy. This optional integration of depth maps allows for flexible adaptation to varying data availability and quality, thereby enhancing the model's performance across different datasets.

\section{Discussion}
\noindent\textbf{Conclusion.} In this study, we introduce FreeSplat++, a generalizable 3DGS model that is tailored to accommodate long sequence of input views and perform free-view synthesis using the global 3D Gaussians. We designed a Low-cost Cross-View Aggregation pipeline and a Pixel-wise Triplet Fusion module, enhancing the model's capability to extend to longer input sequences efficiently.
To address the existing limitations faced by FreeSplat during whole scene reconstruction, FreeSplat++ improved the fusion module to reduce foreground floaters, and designed an effective floater removal module to efficiently remove cross-view floaters, such that we can mitigate the noise in the predicted depth maps. We also use a lower-resolution map for unprojection to improve the efficiency, and designed a depth-regularized fine-tuning process to efficiently optimize the gaussian parameters given the feed-forward 3DGS initialization while maintaining the geometry accuracy.
FreeSplat++ represents a pioneering effort to accomplish the challenging task of feed-forward whole scene reconstruction with 3D Gaussians, and consistently outperforms baseline methods especially when given longer sequence inputs.

\vspace{3pt}
\noindent\textbf{Limitations.} One Limitation of FreeSplat++ is the feed-forward performance of whole scene reconstruction, which is still not comparable to vanilla 3DGS in terms of the color renderings from interpolated views. Consequently, a per-scene fine-tuning step is necessary to enhance the color rendering quality. Future work can explore enhanced fusion module to maintain high rendering quality despite under whole scene reconstruction. Additionally, as shown in Figure \ref{fig:teaser}, FreeSplat++$_{rft}$ still overfits to the training views after fine-tuning, leading to decreased view consistency. Future work can explore integrating extrapolated view consistency constraints into the fine-tuning step to further unleash the view consistency strength of feed-forward 3DGS. 

Overall, FreeSplat++ serves as a pioneering effort in feed-forward whole scene reconstruction with 3DGS, and we hope our work can inspire future research in this direction.

\section{Acknowledgement}
This work is supported by the Agency for Science, Technology and Research (A*STAR) under its MTC Programmatic Funds (Grant No. M23L7b0021).

{
\bibliographystyle{unsrt}
\bibliography{IEEEabrv, main}

\begin{thebibliography}{10}

\bibitem{nerf}
Ben Mildenhall, Pratul Srinivasan, Matthew Tancik, Jonathan Barron, Ravi Ramamoorthi, and Ren Ng.
\newblock Nerf: Representing scenes as neural radiance fields for view synthesis.
\newblock In {\em European Conference on Computer Vision}, pages 405--421. Springer, 2020.

\bibitem{ibrnet}
Qianqian Wang, Zhicheng Wang, Kyle Genova, Pratul~P Srinivasan, Howard Zhou, Jonathan~T Barron, Ricardo Martin-Brualla, Noah Snavely, and Thomas Funkhouser.
\newblock Ibrnet: Learning multi-view image-based rendering.
\newblock In {\em Proceedings of the IEEE/CVF conference on computer vision and pattern recognition}, pages 4690--4699, 2021.

\bibitem{mvsnerf}
Anpei Chen, Zexiang Xu, Fuqiang Zhao, Xiaoshuai Zhang, Fanbo Xiang, Jingyi Yu, and Hao Su.
\newblock Mvsnerf: Fast generalizable radiance field reconstruction from multi-view stereo.
\newblock In {\em Proceedings of the IEEE/CVF International Conference on Computer Vision}, pages 14124--14133, 2021.

\bibitem{mipnerf360}
Jonathan~T Barron, Ben Mildenhall, Dor Verbin, Pratul~P Srinivasan, and Peter Hedman.
\newblock Mip-nerf 360: Unbounded anti-aliased neural radiance fields.
\newblock In {\em Proceedings of the IEEE/CVF Conference on Computer Vision and Pattern Recognition}, pages 5470--5479, 2022.

\bibitem{ngp}
Thomas M{\"u}ller, Alex Evans, Christoph Schied, and Alexander Keller.
\newblock Instant neural graphics primitives with a multiresolution hash encoding.
\newblock {\em ACM transactions on graphics (TOG)}, 41(4):1--15, 2022.

\bibitem{3dgs}
Bernhard Kerbl, Georgios Kopanas, Thomas Leimk{\"u}hler, and George Drettakis.
\newblock 3d gaussian splatting for real-time radiance field rendering.
\newblock {\em ACM Transactions on Graphics}, 42(4):1--14, 2023.

\bibitem{mipsplatting}
Zehao Yu, Anpei Chen, Binbin Huang, Torsten Sattler, and Andreas Geiger.
\newblock Mip-splatting: Alias-free 3d gaussian splatting.
\newblock In {\em Proceedings of the IEEE/CVF Conference on Computer Vision and Pattern Recognition}, pages 19447--19456, 2024.

\bibitem{4dgs}
Guanjun Wu, Taoran Yi, Jiemin Fang, Lingxi Xie, Xiaopeng Zhang, Wei Wei, Wenyu Liu, Qi~Tian, and Xinggang Wang.
\newblock 4d gaussian splatting for real-time dynamic scene rendering.
\newblock In {\em Proceedings of the IEEE/CVF Conference on Computer Vision and Pattern Recognition}, pages 20310--20320, 2024.

\bibitem{neus}
Peng Wang, Lingjie Liu, Yuan Liu, Christian Theobalt, Taku Komura, and Wenping Wang.
\newblock Neus: Learning neural implicit surfaces by volume rendering for multi-view reconstruction.
\newblock {\em arXiv preprint arXiv:2106.10689}, 2021.

\bibitem{neuralangelo}
Zhaoshuo Li, Thomas M{\"u}ller, Alex Evans, Russell~H Taylor, Mathias Unberath, Ming-Yu Liu, and Chen-Hsuan Lin.
\newblock Neuralangelo: High-fidelity neural surface reconstruction.
\newblock In {\em Proceedings of the IEEE/CVF Conference on Computer Vision and Pattern Recognition}, pages 8456--8465, 2023.

\bibitem{neusg}
Hanlin Chen, Chen Li, and Gim~Hee Lee.
\newblock Neusg: Neural implicit surface reconstruction with 3d gaussian splatting guidance.
\newblock {\em arXiv preprint arXiv:2312.00846}, 2023.

\bibitem{vcr}
Hanlin Chen, Fangyin Wei, Chen Li, Tianxin Huang, Yunsong Wang, and Gim~Hee Lee.
\newblock Vcr-gaus: View consistent depth-normal regularizer for gaussian surface reconstruction.
\newblock {\em arXiv preprint arXiv:2406.05774}, 2024.

\bibitem{nesf}
Suhani Vora, Noha Radwan, Klaus Greff, Henning Meyer, Kyle Genova, Mehdi~SM Sajjadi, Etienne Pot, Andrea Tagliasacchi, and Daniel Duckworth.
\newblock Nesf: Neural semantic fields for generalizable semantic segmentation of 3d scenes.
\newblock {\em arXiv preprint arXiv:2111.13260}, 2021.

\bibitem{semanticray}
Fangfu Liu, Chubin Zhang, Yu~Zheng, and Yueqi Duan.
\newblock Semantic ray: Learning a generalizable semantic field with cross-reprojection attention.
\newblock In {\em Proceedings of the IEEE/CVF Conference on Computer Vision and Pattern Recognition}, pages 17386--17396, 2023.

\bibitem{gov}
Yunsong Wang, Hanlin Chen, and Gim~Hee Lee.
\newblock Gov-nesf: Generalizable open-vocabulary neural semantic fields.
\newblock In {\em Proceedings of the IEEE/CVF Conference on Computer Vision and Pattern Recognition}, pages 20443--20453, 2024.

\bibitem{embodied1}
Venkata~Naren Devarakonda, Raktim~Gautam Goswami, Ali~Umut Kaypak, Naman Patel, Rooholla Khorrambakht, Prashanth Krishnamurthy, and Farshad Khorrami.
\newblock Orionnav: Online planning for robot autonomy with context-aware llm and open-vocabulary semantic scene graphs.
\newblock {\em arXiv preprint arXiv:2410.06239}, 2024.

\bibitem{embodied2}
Ri-Zhao Qiu, Yafei Hu, Ge~Yang, Yuchen Song, Yang Fu, Jianglong Ye, Jiteng Mu, Ruihan Yang, Nikolay Atanasov, Sebastian Scherer, et~al.
\newblock Learning generalizable feature fields for mobile manipulation.
\newblock {\em arXiv preprint arXiv:2403.07563}, 2024.

\bibitem{pixelsplat}
David Charatan, Sizhe Li, Andrea Tagliasacchi, and Vincent Sitzmann.
\newblock pixelsplat: 3d gaussian splats from image pairs for scalable generalizable 3d reconstruction.
\newblock {\em arXiv preprint arXiv:2312.12337}, 2023.

\bibitem{mvsplat}
Yuedong Chen, Haofei Xu, Chuanxia Zheng, Bohan Zhuang, Marc Pollefeys, Andreas Geiger, Tat-Jen Cham, and Jianfei Cai.
\newblock Mvsplat: Efficient 3d gaussian splatting from sparse multi-view images.
\newblock {\em arXiv preprint arXiv:2403.14627}, 2024.

\bibitem{gps}
Shunyuan Zheng, Boyao Zhou, Ruizhi Shao, Boning Liu, Shengping Zhang, Liqiang Nie, and Yebin Liu.
\newblock Gps-gaussian: Generalizable pixel-wise 3d gaussian splatting for real-time human novel view synthesis.
\newblock In {\em Proceedings of the IEEE/CVF Conference on Computer Vision and Pattern Recognition}, pages 19680--19690, 2024.

\bibitem{latentsplat}
Christopher Wewer, Kevin Raj, Eddy Ilg, Bernt Schiele, and Jan~Eric Lenssen.
\newblock latentsplat: Autoencoding variational gaussians for fast generalizable 3d reconstruction.
\newblock In {\em European Conference on Computer Vision}, pages 456--473. Springer, 2025.

\bibitem{freesplat}
Yunsong Wang, Tianxin Huang, Hanlin Chen, and Gim~Hee Lee.
\newblock Freesplat: Generalizable 3d gaussian splatting towards free-view synthesis of indoor scenes.
\newblock {\em arXiv preprint arXiv:2405.17958}, 2024.

\bibitem{mvsgaussian}
Tianqi Liu, Guangcong Wang, Shoukang Hu, Liao Shen, Xinyi Ye, Yuhang Zang, Zhiguo Cao, Wei Li, and Ziwei Liu.
\newblock Mvsgaussian: Fast generalizable gaussian splatting reconstruction from multi-view stereo.
\newblock In {\em European Conference on Computer Vision}, pages 37--53. Springer, 2024.

\bibitem{volumetric}
Brian Curless and Marc Levoy.
\newblock A volumetric method for building complex models from range images.
\newblock In {\em Proceedings of the 23rd annual conference on Computer graphics and interactive techniques}, pages 303--312, 1996.

\bibitem{kinectfusion}
Shahram Izadi, David Kim, Otmar Hilliges, David Molyneaux, Richard Newcombe, Pushmeet Kohli, Jamie Shotton, Steve Hodges, Dustin Freeman, Andrew Davison, et~al.
\newblock Kinectfusion: real-time 3d reconstruction and interaction using a moving depth camera.
\newblock In {\em Proceedings of the 24th annual ACM symposium on User interface software and technology}, pages 559--568, 2011.

\bibitem{volume1}
Stephen Lombardi, Tomas Simon, Jason Saragih, Gabriel Schwartz, Andreas Lehrmann, and Yaser Sheikh.
\newblock Neural volumes: Learning dynamic renderable volumes from images.
\newblock {\em arXiv preprint arXiv:1906.07751}, 2019.

\bibitem{volume2}
Vincent Sitzmann, Justus Thies, Felix Heide, Matthias Nie{\ss}ner, Gordon Wetzstein, and Michael Zollhofer.
\newblock Deepvoxels: Learning persistent 3d feature embeddings.
\newblock In {\em Proceedings of the IEEE/CVF Conference on Computer Vision and Pattern Recognition}, pages 2437--2446, 2019.

\bibitem{multiplane}
Tinghui Zhou, Richard Tucker, John Flynn, Graham Fyffe, and Noah Snavely.
\newblock Stereo magnification: Learning view synthesis using multiplane images.
\newblock {\em arXiv preprint arXiv:1805.09817}, 2018.

\bibitem{geogaussian}
Yanyan Li, Chenyu Lyu, Yan Di, Guangyao Zhai, Gim~Hee Lee, and Federico Tombari.
\newblock Geogaussian: Geometry-aware gaussian splatting for scene rendering.
\newblock In {\em European Conference on Computer Vision}, pages 441--457. Springer, 2025.

\bibitem{drgs}
Jaeyoung Chung, Jeongtaek Oh, and Kyoung~Mu Lee.
\newblock Depth-regularized optimization for 3d gaussian splatting in few-shot images.
\newblock In {\em Proceedings of the IEEE/CVF Conference on Computer Vision and Pattern Recognition}, pages 811--820, 2024.

\bibitem{pixelnerf}
Alex Yu, Vickie Ye, Matthew Tancik, and Angjoo Kanazawa.
\newblock pixelnerf: Neural radiance fields from one or few images.
\newblock In {\em Proceedings of the IEEE/CVF Conference on Computer Vision and Pattern Recognition}, pages 4578--4587, 2021.

\bibitem{wang2021ibrnet}
Qianqian Wang, Zhicheng Wang, Kyle Genova, Pratul Srinivasan, Howard Zhou, Jonathan~T. Barron, Ricardo Martin-Brualla, Noah Snavely, and Thomas Funkhouser.
\newblock Ibrnet: Learning multi-view image-based rendering.
\newblock In {\em CVPR}, 2021.

\bibitem{regnerf}
Michael Niemeyer, Jonathan~T Barron, Ben Mildenhall, Mehdi~SM Sajjadi, Andreas Geiger, and Noha Radwan.
\newblock Regnerf: Regularizing neural radiance fields for view synthesis from sparse inputs.
\newblock In {\em Proceedings of the IEEE/CVF Conference on Computer Vision and Pattern Recognition}, pages 5480--5490, 2022.

\bibitem{pointnerf}
Qiangeng Xu, Zexiang Xu, Julien Philip, Sai Bi, Zhixin Shu, Kalyan Sunkavalli, and Ulrich Neumann.
\newblock Point-nerf: Point-based neural radiance fields.
\newblock {\em arXiv preprint arXiv:2201.08845}, 2022.

\bibitem{enerf}
Haotong Lin, Sida Peng, Zhen Xu, Yunzhi Yan, Qing Shuai, Hujun Bao, and Xiaowei Zhou.
\newblock Efficient neural radiance fields for interactive free-viewpoint video.
\newblock In {\em SIGGRAPH Asia 2022 Conference Papers}, pages 1--9, 2022.

\bibitem{transplat}
Chuanrui Zhang, Yingshuang Zou, Zhuoling Li, Minmin Yi, and Haoqian Wang.
\newblock Transplat: Generalizable 3d gaussian splatting from sparse multi-view images with transformers.
\newblock {\em arXiv preprint arXiv:2408.13770}, 2024.

\bibitem{hisplat}
Shengji Tang, Weicai Ye, Peng Ye, Weihao Lin, Yang Zhou, Tao Chen, and Wanli Ouyang.
\newblock Hisplat: Hierarchical 3d gaussian splatting for generalizable sparse-view reconstruction.
\newblock {\em arXiv preprint arXiv:2410.06245}, 2024.

\bibitem{ggrt}
Hao Li, Yuanyuan Gao, Dingwen Zhang, Chenming Wu, Yalun Dai, Chen Zhao, Haocheng Feng, Errui Ding, Jingdong Wang, and Junwei Han.
\newblock Ggrt: Towards generalizable 3d gaussians without pose priors in real-time.
\newblock {\em arXiv preprint arXiv:2403.10147}, 2024.

\bibitem{con}
Songyou Peng, Michael Niemeyer, Lars Mescheder, Marc Pollefeys, and Andreas Geiger.
\newblock Convolutional occupancy networks.
\newblock In {\em Computer Vision--ECCV 2020: 16th European Conference, Glasgow, UK, August 23--28, 2020, Proceedings, Part III 16}, pages 523--540. Springer, 2020.

\bibitem{neuralrecon}
Jiaming Sun, Yiming Xie, Linghao Chen, Xiaowei Zhou, and Hujun Bao.
\newblock Neuralrecon: Real-time coherent 3d reconstruction from monocular video.
\newblock In {\em Proceedings of the IEEE/CVF conference on computer vision and pattern recognition}, pages 15598--15607, 2021.

\bibitem{vortx}
Noah Stier, Alexander Rich, Pradeep Sen, and Tobias H{\"o}llerer.
\newblock Vortx: Volumetric 3d reconstruction with transformers for voxelwise view selection and fusion.
\newblock In {\em 2021 International Conference on 3D Vision (3DV)}, pages 320--330. IEEE, 2021.

\bibitem{simplerecon}
Mohamed Sayed, John Gibson, Jamie Watson, Victor Prisacariu, Michael Firman, and Cl{\'e}ment Godard.
\newblock Simplerecon: 3d reconstruction without 3d convolutions.
\newblock In {\em European Conference on Computer Vision}, pages 1--19. Springer, 2022.

\bibitem{niceslam}
Zihan Zhu, Songyou Peng, Viktor Larsson, Weiwei Xu, Hujun Bao, Zhaopeng Cui, Martin~R Oswald, and Marc Pollefeys.
\newblock Nice-slam: Neural implicit scalable encoding for slam.
\newblock In {\em Proceedings of the IEEE/CVF Conference on Computer Vision and Pattern Recognition}, pages 12786--12796, 2022.

\bibitem{gsslam}
Chi Yan, Delin Qu, Dong Wang, Dan Xu, Zhigang Wang, Bin Zhao, and Xuelong Li.
\newblock Gs-slam: Dense visual slam with 3d gaussian splatting.
\newblock {\em arXiv preprint arXiv:2311.11700}, 2023.

\bibitem{splatam}
Nikhil Keetha, Jay Karhade, Krishna~Murthy Jatavallabhula, Gengshan Yang, Sebastian Scherer, Deva Ramanan, and Jonathon Luiten.
\newblock Splatam: Splat, track \& map 3d gaussians for dense rgb-d slam.
\newblock {\em arXiv preprint arXiv:2312.02126}, 2023.

\bibitem{volsdf}
Lior Yariv, Jiatao Gu, Yoni Kasten, and Yaron Lipman.
\newblock Volume rendering of neural implicit surfaces.
\newblock {\em Advances in Neural Information Processing Systems}, 34:4805--4815, 2021.

\bibitem{monosdf}
Zehao Yu, Songyou Peng, Michael Niemeyer, Torsten Sattler, and Andreas Geiger.
\newblock Monosdf: Exploring monocular geometric cues for neural implicit surface reconstruction.
\newblock {\em Advances in neural information processing systems}, 35:25018--25032, 2022.

\bibitem{gao2023surfelnerf}
Yiming Gao, Yan-Pei Cao, and Ying Shan.
\newblock Surfelnerf: Neural surfel radiance fields for online photorealistic reconstruction of indoor scenes.
\newblock In {\em Proceedings of the IEEE/CVF Conference on Computer Vision and Pattern Recognition}, pages 108--118, 2023.

\bibitem{gaussianroom}
Haodong Xiang, Xinghui Li, Xiansong Lai, Wanting Zhang, Zhichao Liao, Kai Cheng, and Xueping Liu.
\newblock Gaussianroom: Improving 3d gaussian splatting with sdf guidance and monocular cues for indoor scene reconstruction.
\newblock {\em arXiv preprint arXiv:2405.19671}, 2024.

\bibitem{2dgsroom}
Wanting Zhang, Haodong Xiang, Zhichao Liao, Xiansong Lai, Xinghui Li, and Long Zeng.
\newblock 2dgs-room: Seed-guided 2d gaussian splatting with geometric constrains for high-fidelity indoor scene reconstruction.
\newblock {\em arXiv preprint arXiv:2412.03428}, 2024.

\bibitem{vit}
Alexey Dosovitskiy, Lucas Beyer, Alexander Kolesnikov, Dirk Weissenborn, Xiaohua Zhai, Thomas Unterthiner, Mostafa Dehghani, Matthias Minderer, Georg Heigold, Sylvain Gelly, et~al.
\newblock An image is worth 16x16 words: Transformers for image recognition at scale. arxiv 2020.
\newblock {\em arXiv preprint arXiv:2010.11929}, 2010.

\bibitem{swin}
Ze~Liu, Yutong Lin, Yue Cao, Han Hu, Yixuan Wei, Zheng Zhang, Stephen Lin, and Baining Guo.
\newblock Swin transformer: Hierarchical vision transformer using shifted windows.
\newblock In {\em Proceedings of the IEEE/CVF international conference on computer vision}, pages 10012--10022, 2021.

\bibitem{efficientnet}
Mingxing Tan and Quoc Le.
\newblock Efficientnet: Rethinking model scaling for convolutional neural networks.
\newblock In {\em International conference on machine learning}, pages 6105--6114. PMLR, 2019.

\bibitem{resnet}
Kaiming He, Xiangyu Zhang, Shaoqing Ren, and Jian Sun.
\newblock Deep residual learning for image recognition.
\newblock In {\em Proceedings of the IEEE conference on computer vision and pattern recognition}, pages 770--778, 2016.

\bibitem{sweep1}
Robert~T Collins.
\newblock A space-sweep approach to true multi-image matching.
\newblock In {\em Proceedings CVPR IEEE computer society conference on computer vision and pattern recognition}, pages 358--363. Ieee, 1996.

\bibitem{sweep2}
Sunghoon Im, Hae-Gon Jeon, Stephen Lin, and In~So Kweon.
\newblock Dpsnet: End-to-end deep plane sweep stereo.
\newblock {\em arXiv preprint arXiv:1905.00538}, 2019.

\bibitem{deepvideomvs}
Arda Duzceker, Silvano Galliani, Christoph Vogel, Pablo Speciale, Mihai Dusmanu, and Marc Pollefeys.
\newblock Deepvideomvs: Multi-view stereo on video with recurrent spatio-temporal fusion.
\newblock In {\em Proceedings of the IEEE/CVF Conference on Computer Vision and Pattern Recognition}, pages 15324--15333, 2021.

\bibitem{unet++}
Zongwei Zhou, Md~Mahfuzur Rahman~Siddiquee, Nima Tajbakhsh, and Jianming Liang.
\newblock Unet++: A nested u-net architecture for medical image segmentation.
\newblock In {\em Deep Learning in Medical Image Analysis and Multimodal Learning for Clinical Decision Support: 4th International Workshop, DLMIA 2018, and 8th International Workshop, ML-CDS 2018, Held in Conjunction with MICCAI 2018, Granada, Spain, September 20, 2018, Proceedings 4}, pages 3--11. Springer, 2018.

\bibitem{gru}
Rahul Dey and Fathi~M Salem.
\newblock Gate-variants of gated recurrent unit (gru) neural networks.
\newblock In {\em 2017 IEEE 60th international midwest symposium on circuits and systems (MWSCAS)}, pages 1597--1600. IEEE, 2017.

\bibitem{lpips}
Richard Zhang, Phillip Isola, Alexei~A Efros, Eli Shechtman, and Oliver Wang.
\newblock The unreasonable effectiveness of deep features as a perceptual metric.
\newblock In {\em Proceedings of the IEEE conference on computer vision and pattern recognition}, pages 586--595, 2018.

\bibitem{dai2017scannet}
Angela Dai, Angel~X Chang, Manolis Savva, Maciej Halber, Thomas Funkhouser, and Matthias Nie{\ss}ner.
\newblock Scannet: Richly-annotated 3d reconstructions of indoor scenes.
\newblock In {\em Proceedings of the IEEE conference on computer vision and pattern recognition}, pages 5828--5839, 2017.

\bibitem{neuray}
Yuan Liu, Sida Peng, Lingjie Liu, Qianqian Wang, Peng Wang, Christian Theobalt, Xiaowei Zhou, and Wenping Wang.
\newblock Neural rays for occlusion-aware image-based rendering.
\newblock In {\em Proceedings of the IEEE/CVF Conference on Computer Vision and Pattern Recognition}, pages 7824--7833, 2022.

\bibitem{pixelgaussian}
Xin Fei, Wenzhao Zheng, Yueqi Duan, Wei Zhan, Masayoshi Tomizuka, Kurt Keutzer, and Jiwen Lu.
\newblock Pixelgaussian: Generalizable 3d gaussian reconstruction from arbitrary views.
\newblock {\em arXiv preprint arXiv:2410.18979}, 2024.

\bibitem{scannet++}
Chandan Yeshwanth, Yueh-Cheng Liu, Matthias Nie{\ss}ner, and Angela Dai.
\newblock Scannet++: A high-fidelity dataset of 3d indoor scenes.
\newblock In {\em Proceedings of the IEEE/CVF International Conference on Computer Vision}, pages 12--22, 2023.

\bibitem{nerfusion}
Xiaoshuai Zhang, Sai Bi, Kalyan Sunkavalli, Hao Su, and Zexiang Xu.
\newblock Nerfusion: Fusing radiance fields for large-scale scene reconstruction.
\newblock In {\em Proceedings of the IEEE/CVF Conference on Computer Vision and Pattern Recognition}, pages 5449--5458, 2022.

\bibitem{replica}
Julian Straub, Thomas Whelan, Lingni Ma, Yufan Chen, Erik Wijmans, Simon Green, Jakob~J Engel, Raul Mur-Artal, Carl Ren, Shobhit Verma, et~al.
\newblock The replica dataset: A digital replica of indoor spaces.
\newblock {\em arXiv preprint arXiv:1906.05797}, 2019.

\bibitem{semantic_nerf}
Shuaifeng Zhi, Tristan Laidlow, Stefan Leutenegger, and Andrew~J. Davison.
\newblock In-place scene labelling and understanding with implicit scene representation.
\newblock 2021.

\bibitem{adam}
Diederik~P Kingma and Jimmy Ba.
\newblock Adam: A method for stochastic optimization.
\newblock {\em arXiv preprint arXiv:1412.6980}, 2014.

\bibitem{semanticgaussians}
Jun Guo, Xiaojian Ma, Yue Fan, Huaping Liu, and Qing Li.
\newblock Semantic gaussians: Open-vocabulary scene understanding with 3d gaussian splatting.
\newblock {\em arXiv preprint arXiv:2403.15624}, 2024.

\bibitem{nerfrpn}
Benran Hu, Junkai Huang, Yichen Liu, Yu-Wing Tai, and Chi-Keung Tang.
\newblock Nerf-rpn: A general framework for object detection in nerfs.
\newblock In {\em Proceedings of the IEEE/CVF conference on computer vision and pattern recognition}, pages 23528--23538, 2023.

\bibitem{roessle2022dense}
Barbara Roessle, Jonathan~T Barron, Ben Mildenhall, Pratul~P Srinivasan, and Matthias Nie{\ss}ner.
\newblock Dense depth priors for neural radiance fields from sparse input views.
\newblock In {\em Proceedings of the IEEE/CVF Conference on Computer Vision and Pattern Recognition}, pages 12892--12901, 2022.

\bibitem{ssim}
Zhou Wang, Alan~C Bovik, Hamid~R Sheikh, and Eero~P Simoncelli.
\newblock Image quality assessment: from error visibility to structural similarity.
\newblock {\em IEEE transactions on image processing}, 13(4):600--612, 2004.

\end{thebibliography}
}

\vfill

\end{document}